\pdfoutput=1

\documentclass[11pt]{article}

\usepackage[final]{acl}

\usepackage{times}
\usepackage{latexsym}

\usepackage[T1]{fontenc}

\usepackage[utf8]{inputenc}

\usepackage{microtype}

\usepackage{inconsolata}

\usepackage{graphicx}
\usepackage[frozencache,cachedir=.]{minted} 
\usepackage[many]{tcolorbox}  
\newtcolorbox{boxC}{
    boxrule = 0pt,
    breakable,
}
\newtcolorbox{boxL}{
    fontupper = \color{black},
    rounded corners,
    arc = 6pt,
    colframe = black!50, 
    boxrule = 0pt, 
    bottomrule = 4.5pt ,
    breakable,
}
\usepackage{tabularx, colortbl,booktabs}
\usepackage{soul}
\newcolumntype{Y}{>{\centering\arraybackslash}X}
\usepackage{hyperref}

\title{Large Language Models Can Solve Real-World Planning Rigorously\\ with Formal Verification Tools}

\author{Yilun Hao \\
  MIT \\
  \texttt{yilunhao@mit.edu} \\\And
  Yongchao Chen \\
  MIT / Harvard University \\
  \texttt{yongchaochen@fas.harvard.edu} \\\AND
  Yang Zhang \\
  MIT-IBM Watson AI Lab \\
  \texttt{Yang.Zhang2@ibm.com}\\\And
  Chuchu Fan \\
  MIT\\
  \texttt{chuchu@mit.edu}\\}

\usepackage{color}

\begin{document}
\maketitle
\begin{abstract}

Large Language Models (LLMs) struggle to directly generate correct plans for complex multi-constraint planning problems, even with self-verification and self-critique.
For example, a U.S. domestic travel planning benchmark TravelPlanner was proposed in~\citet{xie2024travelplanner}, where the best LLM OpenAI o1-preview can only find viable travel plans with a 10\% success rate given all needed information. In this work, we tackle this by proposing an LLM-based planning framework that formalizes and solves complex multi-constraint planning problems as constrained satisfiability problems, which are further consumed by sound and complete satisfiability solvers. 
We start with TravelPlanner as the primary use case and show that our framework achieves a success rate of 93.9\% and is effective with diverse paraphrased prompts.
More importantly, our framework has strong zero-shot generalizability, successfully handling unseen constraints in our newly created unseen international travel dataset and generalizing well to new fundamentally different domains.  
Moreover, when user input queries are infeasible, our framework can identify the unsatisfiable core, provide failure reasons, and offers personalized modification suggestions. We show that our framework can modify and solve for an average of 81.6\% and 91.7\% unsatisfiable queries from two datasets and prove with ablations that all key components of our framework are effective and necessary. Project page: \href{https://sites.google.com/view/llm-rwplanning}{https://sites.google.com/view/llm-rwplanning}.
\end{abstract}
\section{Introduction}
Recent work has demonstrated that large language models (LLMs) \cite{brown2020language, ouyang2022training, achiam2023gpt}, with abundant world knowledge, abilities to collect information via tools, and capabilities of reasoning, have significant potential in solving planning problems~\cite{huang2022language, ahn2022can, yao2022react, huang2022inner, song2023llm}. 
However, modern LLMs are not well-suited for directly solving highly complex combinatorial optimization problems with multiple levels of constraints. This is because LLMs generate responses based on token probabilities derived from their training data and do not inherently possess the ability to perform rigorous logical or mathematical reasoning.
To investigate the performance of LLMs on complex realistic multi-constraint problems, \citet{xie2024travelplanner} proposed a U.S. domestic travel planning benchmark, TravelPlanner, and showed that LLMs are not capable of handling this task and the best LLM at the time, GPT-4, only achieves a 0.6\% success rate. We test the strongest model of today, OpenAI o1-preview~\cite{o1preview}, with TravelPlanner and observed a pass rate of 10.0\% even with access to pre-collected information.
LLM-Modulo Framework~\cite{kambhampati2024llms}, a recent work that combines LLMs with external critics, verifiers, and humans, raises the pass rate to 20\% with GPT-4-Turbo and 65\% with o1-preview, which is the best performance on TravelPlanner as of now.

To tackle multi-constraint problems like travel planning, an alternative way is through constraint-based planning to formalize the problem as a constraint satisfaction problem (CSP)~\cite{dechter2003constraint, lozano2014constraint}, including boolean satisfiability problem (SAT)~\cite{kautz1999unifying, rintanen2012planning} and satisfiability modulo theory (SMT)~\cite{barrett2010smt, de2011satisfiability, dantam2016incremental}, and solve it with existing algorithm-based solvers~\cite{dutertre2006fast, de2008z3, barrett2011cvc4}. However, algorithm-based solvers usually have steep learning curves. As natural language queries have no fixed format, planners need to extract key information from input queries accurately to model the problem.
Crucially, even if the extracted information is correct, users must still modify inputs and query the tools repeatedly if their inputs are unsatisfiable.

In short, LLM-based and algorithm-based planning methods have complementary strengths: LLMs excel at parsing human input and interactions but struggle to rigorously solve complex planning problems with multiple constraints. In contrast, algorithm-based solvers are sound and complete when solving multi-constraint satisfiability problems but are incapable of handling dynamic, general, and sometimes ambiguous natural language requirements. Can we design a framework that combines the merits of both paradigms and enables a strong, rigorous, and yet user-friendly planning experience for human users?

Motivated by this, in this paper, we propose a travel-planning framework that enables LLMs to process human queries and generate code to automatically utilize algorithm-based solvers, \emph{e.g.}, the SMT solver, to formally formulate, solve, and reason over the planning problem. Specifically, we take travel planning as the primary use case and provide LLMs with instruction steps and corresponding codes for using SMT solver to solve the example travel planning problems. We find that, with only three examples in the prompt, the LLM can effectively learn the pattern and generalize to new input queries. Since a solver is called to solve the problem with all the constraints encoded, our method is guaranteed to generate a plan if it exists.

Notably, our method offers following features, including and beyond aforementioned advantages.

\noindent $\bullet$ \textbf{Superior Planning Success Rate.} We evaluate our framework over different LLMs, GPT-4, Claude 3 Opus, and Mistral-large, and show that our framework achieves the best final pass rates of 93.3\% and 93.9\% on TravelPlanner validation and test sets. Our framework significantly outperforms the best tool-use framework on TravelPlanner, LLM Modulo, which achieves pass rates of 20\% and 65\% with GPT-4-Turbo and o1-preview.

\noindent $\bullet$ \textbf{Feedback Interactions on Unsolvable Cases.}  If the input query is not satisfiable, our framework utilizes SMT solvers to identify the exact unsatisfiable constraints, analyzes the unsatisfiable reasons, and proposes suggestions to modify the query until it becomes satisfiable. In addition to a fully autonomous mode of offering default modification suggestions by itself, LLM can even interactively communicate with humans to incorporate their unique preferences. Our experiments show that our feedback mechanism can turn an average of 81.6\% and 91.7\% insolvable problems from two datasets into solvable ones under different user reaction patterns to our model's suggestions.

\noindent $\bullet$ \textbf{Strong Zero-Shot Generalizability to Unseen Constraints and Tasks.} Although the proposed planning system is demonstrated with only travel planning examples with a fixed set of constraints, we found that it is readily generalizable to unseen constraints and even unseen planning tasks. Remarkably, we further introduce four completely new domains that involve combinatorial tasks: Block Picking, Task Allocation, Travelling Salesman Problem, and Warehouse. On all new scenarios, our framework achieves an average of 89.0\% optimal rates for four new domains in a zero-shot manner, as discussed in Section~\ref{sec:gen-task}.

\noindent $\bullet$ \textbf{High Prompt Robustness.} To address concerns that the strong planning capabilities of our system might result from extensive prompt engineering, we conduct experiments to evaluate its robustness to varied paraphrased prompts. Our results show that the system effectively handles a wide range of prompt formulations, thus is not overly reliant on specific prompt designs. This highlights the flexibility and robustness of our method, establishing it as a powerful and adaptable planning paradigm.


\begin{figure*}[!ht]
  \includegraphics[width=\linewidth]{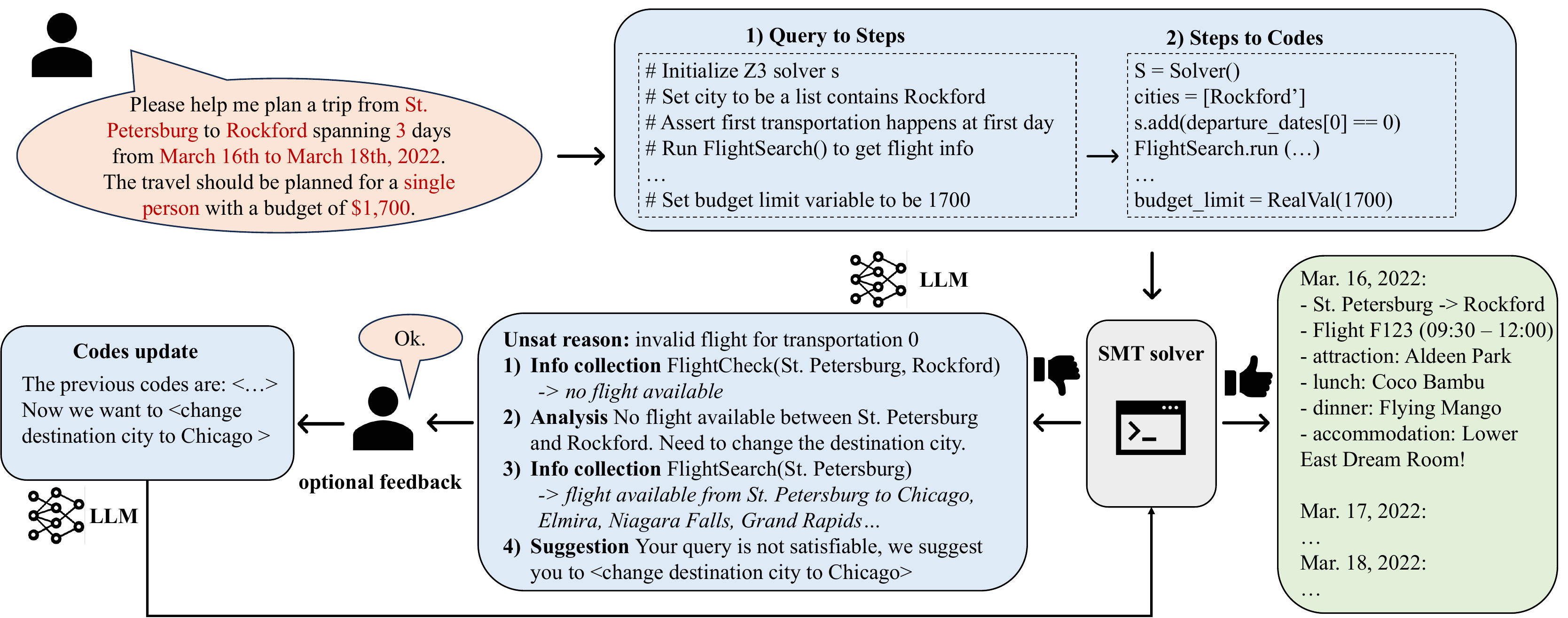} 
  \caption {An overview of the framework. The blue region represents LLM. Given a natural language query, LLM 1) generates steps to formulate it as an SMT problem, 2) generates corresponding codes that encode the problem and call the solver. If the solver is not able to find the solution, LLM receives unsatisfiable reasons from the solver, collects information, analyzes the current situation, and offers suggestions to modify the query interactively. LLM then updates the code based on suggestions and calls the solver again to find a feasible plan.}
  \label{fig:Framework}
  \vspace{-12pt}
\end{figure*}
\section{Related Work}

\noindent \textbf{LLM Planning.} 
LLMs have shown significant intelligence in reasoning~\cite{wei2022chain, kojima2022large, yao2022react} and tool-use~\cite{qin2023toolllm, schick2024toolformer}, offering the potential of promising planning capability. Previous works tackle planning problems with various ways~\cite{huang2024understanding}: 1) decomposing the task into sub-tasks~\cite{wei2022chain, yao2022react, shen2024hugginggpt}; 2) generating multiple plans and selecting the optimal one~\cite{wang2022self, yao2024tree, zhao2024large, besta2024graph, hao2023reasoning}; 3) reflecting on experiences and refining plan based on feedback~\cite{shinn2024reflexion, madaan2024self,chen2023scalable}; 4) formalizing tasks and aiding with external planner~\cite{liu2023llm+, guan2023leveraging,chen2023autotamp}. While these planning algorithms have shown promising results, their planning scenarios are limited to simple tasks with a limited number of constraints. \citet{xie2024travelplanner} proposes a realistic complex travel planning benchmark and tests on various LLM planning methods to show that LLMs are not capable of handling multi-constraint tasks. 

\noindent \textbf{Algorithm-based Planning.}
Another way to tackle travel planning is through algorithm-based planning such as heuristic search~\cite{ hoffmann2001ff, helmert2006fast,vidal2014yahsp3} and constraint-based methods~\cite{kautz1999unifying, rintanen2012planning, rintanen2014madagascar,lozano2014constraint,dantam2016incremental}. However, these methods can not generalize to diverse natural language inputs and may not guarantee to find the plan. Our framework enables LLM to utilize constraint-based planning methods by formalizing diverse human queries into an SMT problem and solving with sound and complete SMT solvers. 

\noindent \textbf{LLM Tool-use.}
Tool-using allows LLMs to utilize powerful external tools to increase reliability. Recent works explore how LLMs could utilize external tools such as search engines, operating environments, and code generators~\cite{press2022measuring, yao2022react, schick2024toolformer, liang2023code, singh2023progprompt, peng2023check,song2023llm, huang2022inner, yuan2024easytool} to provide feedback or extra information. In our framework, LLMs generate codes to formulate the travel planning problem as an SMT problem and call the SMT solver. This overcomes LLM's failure to consider all constraints by encoding and solving all constraints rigorously. 

\noindent \textbf{LLM Prompt Design.}
\label{sec:related-prompt}
Prompt design is critical for LLM-based agents to improve the performance of black-box LLMs. While many works automatically evolve prompts through iterations~\cite{wang2023promptagent, fernando2023promptbreeder, chen2024prompt}, in-context learning that includes example input-output pairs in prompts is a cheaper, widely-adopted, and reliable way~\citep{brown2020language, wei2022chain, liang2023code}. Although designing in-context examples requires task-specific efforts, most existing methods proposing to solve complex planning problems require different forms of task-specific efforts~\citep{liu2023llm+, li2023large, gundawar2024robust}.
While our work requires prompt design, these designs are offline and not needed for end users. More importantly, we have shown that the performance of our framework is not sensitive to the specific wording of prompts (Section~\ref{sec:gen-prompt}). Meanwhile, our framework generalizes and achieves great performance for other constraints and even in other tasks without the specific design of prompts (Sections~\ref{sec:gen-constraint} and~\ref{sec:gen-task}).

\section{Approach}

\subsection{Problem Formulation}

In the framework design, we primarily focus on the specific travel planning task with a set of pre-defined constraints. However, we will show that the same prompts and workflow can readily be applied to new constraints and other planning scenarios.

Our travel planning problem is formulated as follows. Given a natural language description of humans' constraints $\mathcal C$ of a travel plan, the system should output a plan that satisfies $\mathcal C$. The travel starts from city $o$, travels $k$ destination cities, and returns to $o$. The travel spans $n$ days. The travel takes $k+1$ transportation methods for $k+1$ travels from city to city. The travelers visit $x$ attractions, dine in $y$ restaurants, and live in accommodations for $n-1$ nights. By default, we set $x=n$, $y=3n$. However, this is not a fixed requirement. Users could specify their unique requirements by adding descriptions in prompts, for example, \textit{``Number of attractions to visit per day is 2''}. Table~\ref{constraint} summarizes the constraints $\mathcal C$ for two datasets we used. The output plan should satisfy $\mathcal C$ and specify the city to visit, transportation method, attraction, restaurant, and accommodation for each day. See Appendix~\ref{sec:appendix-ex} for example input query and output plan.

\subsection{Framework Overview}

As shown in Fig.~\ref{fig:Framework}, if the planning problem comes with valid solutions, our framework solves the problem in four steps. First, an LLM is prompted to parse the user request and output a set of steps to convert the user descriptions to a formal planning problem. Second, the LLM is prompted to further convert the steps to code. Third, the framework calls an external formal solver, in our case the SMT solver, to execute the code, whose output is then parsed by the LLM into natural language outputs.

However, there are cases where the user queries do not bear valid solutions, \emph{e.g.}, the users request to stay in five-star hotels but the specified budget is too low, where the aforementioned steps would fail to generate a valid plan. In these cases, the LLM is prompted to reason about the situations to give suggestions to modify the constraints, \emph{e.g.}, increasing the budget. If users accept the suggestions, the modified planning problem is formed and sent to the solver; otherwise, new suggestions will be made until an agreement is reached or timeout.

Section~\ref{subsec:satisfiable} details the three steps to solve satisfiable planning problems; Section~\ref{sec:planrepair} describes our approach to repairing unsatisfiable plans iteratively. All the prompts are listed in Appendix~\ref{sec:appendix-prompt}.

\subsection{Satisfiable Plan Solving}
\label{subsec:satisfiable}

\subsubsection{Query-Step Generation}
\label{subsubsec:query-step}

Query-Step generation involves transforming natural language queries into a sequence of executable steps, expressed in natural language, to formulate the constraints. For example, to specify the \textit{``travel spanning 3 days''} constraint in the query in Fig.~\ref{fig:Framework}, an example sequence of steps is:
\vspace{-2pt}
\begin{minted}[fontsize=\scriptsize]{text}
1. Set 't_dates' variables for 2 transportation between cities
2. Assert first transportation happens at first day (day 0), 
   and last happens at last day (day 2)
\end{minted}
\vspace{-6pt}
For the query \textit{``travel 2 destination cities in 5 days''}, an example sequence of steps is: 
\vspace{-2pt}
\begin{minted}[fontsize=\scriptsize]{text}
1. Set 't_dates' variables for 3 transportation between cities
2. Assert first transportation happens at first day (day 0), 
   last happens at last day (day 4),
   and second could happen at any day in between
\end{minted}
\vspace{-6pt}

We teach the LLM to perform such generation by providing three human-crafted examples, each containing a natural language query and the corresponding steps. Since the steps to formulate different constraints are different, the steps in each example are broken into nine sections, each focusing on the constraints about one particular subject, such as destinations cities, departure dates, transportation methods, \emph{etc.} (See Appendix~\ref{sec:appendix-prompt} for the detailed prompt). It is important to note that although in-context examples only concern a limited number of constraints, the LLM is able to learn generic patterns and generalize to unseen constraints.

Optionally, converting natural language queries into fixed format JSON descriptions before generating the encoding steps could help LLMs summarize the key information and further improve the performance. However, prompting LLMs to generate JSON descriptions requires explanations of the needed fields in JSON, which could expose more information that helps LLMs to better understand the problem. For fairness, we do not generate JSON files when compared with other methods with natural language inputs. We include the result of our framework with this extra step in Appendix~\ref{sec:appendixcours+json}.
\begin{figure}[t]
\includegraphics[width=\linewidth]{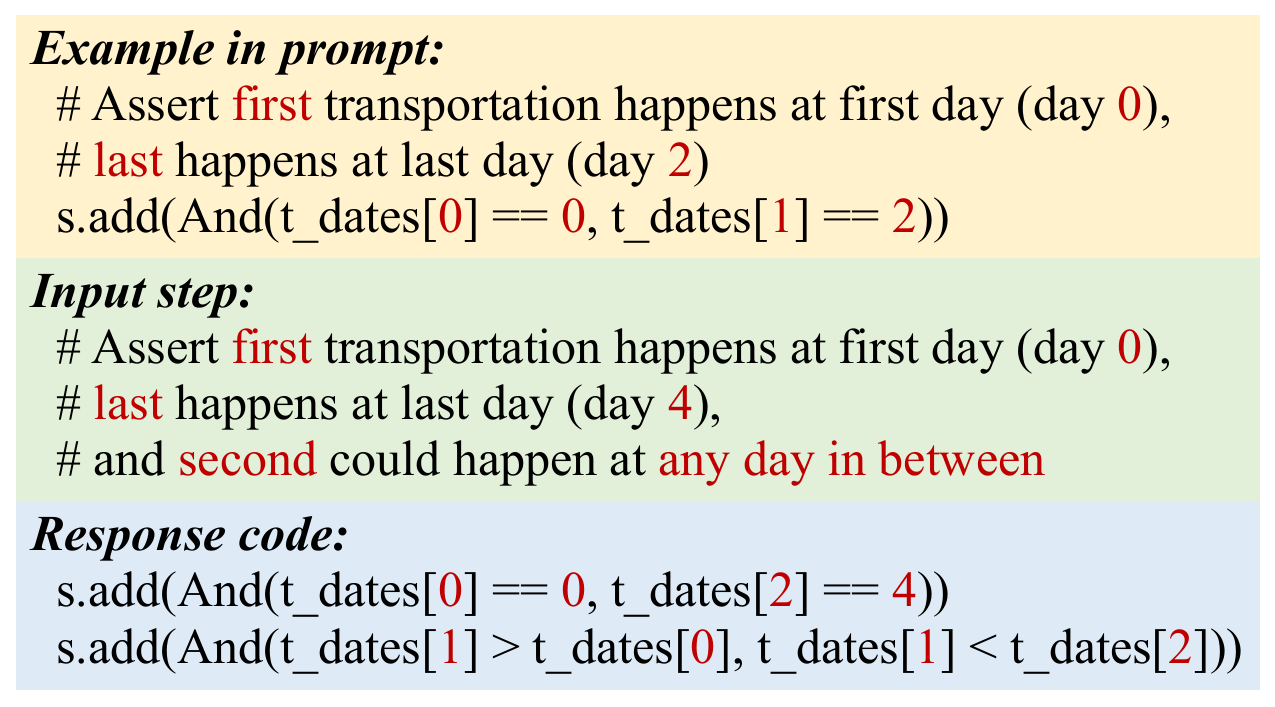} 
\vspace{-20pt}
  \caption {Step to Code translation example.}
  \label{fig:code}
  \vspace{-15pt}
\end{figure}
\subsubsection{Step-Code Generation}
Step-code generation involves converting each of the steps generated in the previous stage into Python code, which would call the relevant APIs that collect information, such as \texttt{CitySearch}, \texttt{FlightSearch}, \texttt{AttractionSearch} \emph{etc.}, as well as the SMT solver to execute the steps. Inspired by \citet{liang2023code}, we teach the LLM to generate the code using examples as demonstrations, as shown in Fig.~\ref{fig:Framework} (part 3). Each example contains a single step drawn from the query-code generation examples (Section~\ref{subsubsec:query-step}), and the corresponding human-written code. To ensure comprehensiveness, our step-code examples cover almost all the steps mentioned in the query-step examples, less some duplicated or highly similar steps. Since the examples already contain sufficient demonstrations of all the API and SMT solver calls, the LLM can learn to use these APIs and the solver without separate API documentation. Fig.~\ref{fig:code} shows how the LLM generalizes to new instruction steps to write corresponding codes given these examples.

\subsubsection{SMT Solver}
After generating codes, our framework executes these codes to encode the problem and call the SMT solver. Since the SMT solver is sound and complete, it guarantees to find a solution if there exists one. Thus, if the constraints are satisfiable, the solver generates a formally verified plan. If the constraints are not satisfiable, the solver outputs the unsatisfiable reasons and LLM could, based on its commonsense and reasoning capabilities, analyze the reasons, actively collect more information, and provide humans with suggestions to modify the constraints. We extract the unsatisfiable reasons with Z3 solver's \verb|get_unsat_core| function. When the framework proves the constraints to be unsatisfiable, it proceeds to interactive plan repair with the unsatisfiable reasons. 

\subsection{Interactive Plan Repair}
\label{sec:planrepair}
When a proposed query is not satisfiable, LLM's reasoning capability and commonsense knowledge to analyze the current situation and offer suggestions become vital. Furthermore, these capabilities enable an interactive setting, in which humans can agree, disagree, or provide feedback to LLM's proposed suggestions. LLM can deliver personalized plans built upon different human preferences.

Inspired by ReAct~\cite{yao2022react}, in our framework, LLM could either take an action to collect information based on unsatisfiable reasons, analyze the current situation based on collected information, or provide suggestions. We equip LLM with information collection APIs and descriptions of their usage. As shown in Fig.~\ref{fig:Framework}, the unsatisfiable reason is \textit{``invalid flight for transportation 0''}. With the reason, the LLM first collects flight information with FlightCheck API. Realizing no flight is available between St. Petersburg and Rockford, LLM analyzes and decides to change the destination city. It then runs FlightSearch API to search for all available destinations and chooses one of them. LLM offers this as a suggestion to the user. The user can optionally provide feedback, including yes, no, any natural language preference, or even modifications users proposed. If the user disagrees with the suggestion or provides their preferences, the framework starts another iteration and proposes new suggestions. If the users do not provide feedback, agree with the suggestion, or propose their own modification, the framework continues by inputting this modification, together with original codes, to an LLM and prompting it to modify the codes. By running the modified codes, the framework generates a plan if the modified constraints are satisfiable. Otherwise, the framework will gather the unsatisfiable reasons and start another round. 
\section{Dataset}
To access our framework's ability to 1) generalize to unseen constraints and 2) interactive plan repair for unsatisfiable queries, we propose a dataset, UnsatChristmas, that introduces new constraints not included in TravelPlanner and contains 39 unsatisfiable queries under this setting. The queries in UnsatChristmas aim to create an international travel plan for Christmas week in 2023. We set cities in our dataset to be the top ten worldwide city destinations in 2019\footnote{\url{https://go.euromonitor.com/white-paper-travel-2019-100-cities.html}} and obtain attraction information from Metabase\footnote{\url{https://www.metabase.com/blog/data-guide-to-travel}}. We utilize Google Flights\footnote{\url{https://www.google.com/travel/flights}} to collect flight information from 12-24-2023 to 12-30-2023 for these ten cities. Compared with TravelPlanner, we omit detailed information on transportation methods, restaurants, and accommodations but introduce detailed constraints regarding flights and attractions. As shown in Table~\ref{constraint}, UnsatChristmas allows users to specify preferences for 1) non-stop flights, 2) the list of airlines, and 3) the list of attraction categories. We collect 39 unsatisfiable queries with 4 possible reasons: non-stop flight rule not satisfied, flight airline requirement not satisfied, attraction category requirement not satisfied, budget not enough. Out of the 39 queries, 12 fail due to one reason, 18 fail due to two reasons, 8 fail due to three reasons, and 1 fails due to four reasons. There are 13 queries with a single destination city, 13 with two, and 13 with three. In addition, to test the interactive plan repair performance, we also modify 12 queries from the training set of TravelPlanner to be unsatisfiable.

\section{Experimental Results}
\begin{table*}[!ht]
\caption{
    Performance comparison of satisfiable queries for 180 queries in the validation set and 1000 queries in test set of TravelPlanner. The results of Greedy Search, TwoStage, and Direct with GPT-4 are from~\citet{xie2024travelplanner}
}
\label{sat_result}
\vspace{-8pt}
\begin{small}
\begin{tabularx}{\textwidth}{l *{6}{Y}}
\toprule
 & Delivery & \multicolumn{2}{c}{Commonsense}
 & \multicolumn{2}{c}{Hard Constraint} & Final\\
 Method & Rate & \multicolumn{2}{c}{Pass Rate}
 & \multicolumn{2}{c}{Pass Rate} & Pass Rate\\
\cmidrule(lr){3-4} \cmidrule(l){5-6}
  &  & Micro & Macro & Micro & Macro & \\
\midrule
 \rowcolor{lightgray}\multicolumn{7}{c}{\textit{\textbf{Validation} (\#180)}}\\
 \midrule
 Greedy Search & \textbf{100} & 74.4 & 0 & 60.8 & 37.8 & 0\\
 TwoStage (GPT-4)  &  89.4 & 61.1 &  2.8 &  15.2 & 10.6 &  0.6\\
 Direct (GPT-4)  &  \textbf{100} & 80.4 &  17.2 &  47.1 & 22.2 &  4.4\\
 Direct (o1-preview)  &  \textbf{100} & 79.6 &  15.0 &  41.9 & 37.8 &  10.0\\
Ours (Mistral-Large)  &  72.2 & 72.0 & 70.6 & 63.3 & 66.7 & 66.7\\
Ours (Claude-3)  &  96.1 & \textbf{96.0} & \textbf{95.6} & 94.8 & 93.3 & \textbf{93.3} \\
  Ours (GPT-4)  &  95.0 & 95.0 & 95.0  &  \textbf{95.7} &  \textbf{98.9} & \textbf{93.3}\\
\midrule
 \rowcolor{lightgray}\multicolumn{7}{c}{\textit{\textbf{Test} (\#1000)}}\\
 \midrule
  Greedy Search & \textbf{100} & 72.0 & 0 & 52.4 & 31.8 & 0\\
 TwoStage (GPT-4)  &  93.1 & 63.3 &  2.0 &  10.5 & 5.5 &  0.6\\
 Direct (GPT-4)  &  \textbf{100} & 80.6 &  15.2 &  44.3 & 23.1 &  4.4\\
 Ours (Mistral-Large)  & 69.9  & 69.8   & 69.4   & 63.0  & 67.8 & 67.8\\
Ours (Claude-3)  & 95.4  & \textbf{95.2} & \textbf{94.3}  &  \textbf{93.5} &  \textbf{93.9} & \textbf{93.9}\\
  Ours (GPT-4)  &  91.5 & 91.4 & 91.1  & 91.3 &  90.2 & 90.2\\
\bottomrule
\end{tabularx}
\end{small}
\vspace{-6pt}
\end{table*}
We examine our framework on both TravelPlanner and UnsatChristmas. We use GPT-4~\cite{achiam2023gpt} with temperature 0 by default, and we also compare with Claude 3 Opus-20240229~\cite{claude} and Mistral-Large~\cite{Mistral} with temperature 0 for satisfiable plan solving evaluation.  We use Z3 SMT solver~\cite{de2008z3}. Since the solution space is prohibitive considering the combinatorial choices and a few queries have very limited feasible plans, we limit the SMT solver's maximum runtime for each query to 30 minutes. Please refer to Appendix~\ref{sec:appendix-time} for cost and runtime analysis.
\subsection{Satisfiable Plan Solving Evaluation}
We examine how well our framework can create travel plans for satisfiable natural language queries on the TravelPlanner benchmark. We design our example instruction steps and corresponding codes using three queries from TravelPlanner's training set and tune the prompt with other queries in the training set. We evaluate our method on both the validation (180 queries) and the test set (1000 queries). \\
\textbf{Evaluation Metric}\quad 
We adopt evaluation metrics from~\citet{xie2024travelplanner} and mainly look at the Final Pass Rate, which represents whether LLMs pass all constraints. Please refer to Appendix~\ref{sec:appendixctype} for detailed descriptions of other evaluation metrics.
\\
\textbf{Baselines}\quad
We compare our framework with the three strongest models using different strategies from \citet{xie2024travelplanner}. Greedy Search uses a traditional search algorithm and heuristically optimizes for total cost. TwoStage (GPT-4), a two-stage tool-use framework, collects information with ReAct~\cite{yao2022react} and then gives plans. Direct (GPT-4), a sole-planning framework, has access to all necessary information and gives plans without tool-calling needs. We also include the result of Direct (o1-preview), the strongest reasoning model so far, for the validation set. Due to the long runtime of o1-preview, we do not evaluate Two-Stage or our framework with o1-preview. To verify the effectiveness of our framework in varied LLMs, we also evaluate with Claude 3 Opus and Mistral-Large. We tune the prompt with the training set and include the prompt differences in Appendix~\ref{sec:appendix-sat4}.\\
\textbf{Results and Analysis}\quad
Table~\ref{sat_result} shows the performance comparison. 
From the results, all LLM planning methods, TwoStage (GPT-4), Direct (GPT-4), and Direct (o1-preview), struggle to take all constraints into consideration with a final pass rate of 0.6\%, 4.4\%, and 10.0\%. Without formal specification, Greedy Search fails to pass any of the tasks. Ours (Claude-3), with the capability of formally encoding the problem as an SMT problem, achieves the highest final pass rate of 93.3\% and 93.9\% for validation and test set. This demonstrates our framework's robustness in solving satisfiable queries. In addition, Ours (GPT-4) could reach comparable results as Ours (Claude-3). Although the pass rate for Ours (Mistral-Large) drops 26.6\% and 26.1\% compared to Ours (Claude-3), it still significantly outperforms baselines, and 92.3\% and 97.0\% of its delivered plans are correct plans. See Appendix~\ref{sec:appendix-sat5} for major failure cases of Ours (Mistral-Large). These results demonstrate the adaptability of our framework to various LLMs. 

In addition, since inputs of baselines are natural language only, to ensure fairness, we do not include the additional step to translate natural language inputs into JSON representations. However, we show that including this step could even further improve the performance by helping LLMs extract and summarize key information before generating steps. We test on all three LLMs and achieve 98.9\%, 98.3\%, and 84.4\% pass rates with on average 9.4\% improvements on the validation set. We also test GPT-4 on the test set, which achieves 97.0\%. Please refer to Appendix~\ref{sec:appendixcours+json} for details.
\subsection{Generalization Capability Analysis}
\begin{table*}[!ht]
\caption{
    Performance of zero-shot generalization to four other combinatorial optimization tasks.
}
\label{newtasks}
\vspace{-8pt}
\begin{small}
\begin{tabularx}{\textwidth}{l *{8}{Y}}
\toprule
 & \multicolumn{2}{c}{Block Picking}
 & \multicolumn{2}{c}{Task Allocation}
 & \multicolumn{2}{c}{TSP}
 & \multicolumn{2}{c}{Warehouse}\\
 \cmidrule(lr){2-3}  \cmidrule(lr){4-5} \cmidrule(l){6-7} \cmidrule(l){8-9}
 Method & Delivery & Optimal & Delivery & Optimal & Delivery & Optimal & Delivery & Optimal \\
 \midrule
TwoStage(GPT-4o)  & 80 & 4 & 84 & 0 & \textbf{100} & 0 &  72 & 0 \\
Ours(GPT-4o)  & \textbf{100} & \textbf{92} & \textbf{92} & \textbf{92} & \textbf{100} & \textbf{100} & \textbf{84}  & \textbf{72} \\
\bottomrule
\end{tabularx}
\end{small}
\vspace{-6pt}
\end{table*}
\begin{table*}[!ht]
\caption{
    Performance of interactive plan repair for unsatisfiable queries on 39 queries from UnsatChristmas.
}
\label{unsat_result_1}
\vspace{-8pt}
\begin{small}
\begin{tabularx}{\textwidth}{l *{6}{Y}|Y}
\toprule
 Method & Always Agree & Budget & Non-stop & Airline & Attraction Category & Destination Cities & Average\\
\midrule
No Reason  &  74.4 & \textbf{61.5} & 69.2  &  53.8 &  69.2 & 53.8 & 63.7\\
No Feedback  &  N/A & 59.0 & 79.5  &  61.5 &  79.5 & 74.4 & 70.8\\
No Solver  &  25.6 & 20.5 & 28.2 & 20.5  & 23.1  & 33.3 & 25.2\\
Ours  &  89.7 & 59.0 & 84.6  &  64.1 &  89.7 & 84.6 & 78.6\\
Ours-20  & \textbf{92.3}  & \textbf{61.5} & \textbf{87.2} & \textbf{66.7} & \textbf{89.7} & \textbf{92.3} & \textbf{81.6}\\
\bottomrule
\end{tabularx}
\end{small}
\vspace{-6pt}
\end{table*}

\begin{table*}[!ht]
\caption{
    Performance of interactive plan repair for unsatisfiable queries on 12 modified queries from TravelPlanner.
}
\label{unsat_result_2}
\vspace{-8pt}
\begin{small}
\begin{tabularx}{\textwidth}{l *{5}{Y}|Y}
\toprule
 Method & Always Agree & Budget & Destination Cities & Transportation Methods & House Type & Average\\
\midrule
No Reason  &  75 & \textbf{83.3} & 91.7 & 83.3 & 66.7 & 80\\
No Feedback  &  N/A & 50 & 91.7 & 66.7 & 75 & 70.9\\
No Solver  & 16.7  & 16.7 & 50  & 25  &  16.7 & 25.0\\
Ours  &  91.7 & 75 & \textbf{100} & 83.3 & 75 & 85.0\\
Ours-20  &  \textbf{100} & \textbf{83.3} & \textbf{100} & \textbf{91.7} & \textbf{83.3} & \textbf{91.7}\\
\bottomrule
\end{tabularx}
\end{small}
\vspace{-6pt}
\end{table*}
\begin{figure}[t]
  \includegraphics[width=\linewidth]{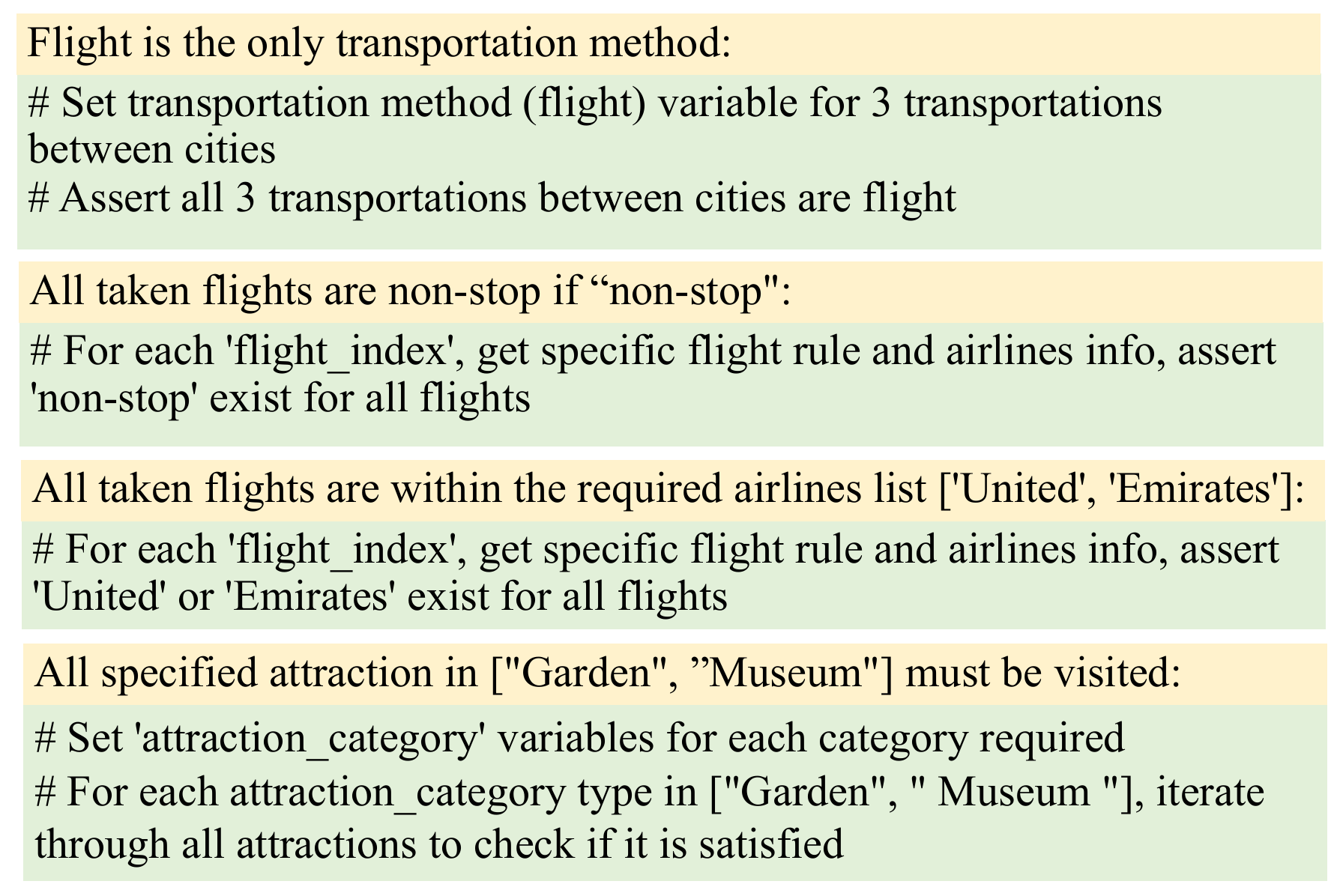} 
  \vspace{-20pt}
  \caption {Example of how JSON-Step prompt generalizes to unseen constraints. Yellow: unseen constraint types. Green: corresponding generated steps.}
  \label{fig:generalization}
\vspace{-18pt}
\end{figure}
\subsubsection{New Prompts}
\label{sec:gen-prompt}
Since both Query-Step and Step-Code prompts require careful design, we test the compatibility of our framework to handle diverse new prompts. We use GPT-4o~\cite{gpt-4o} to paraphrase the entire Query-Step and Step-Code prompts, including the examples used in prompts, for every query in TravelPlanner's validation set and repeat for all 180 queries. We use the paraphrased prompts to test our framework, which achieves a final pass rate of 86.7\% for GPT-4. This showcases the outstanding performance of our framework is not sensitive to the wording of prompts and does not heavily rely on the prompt design. 
Please refer to Appendix~\ref{sec:appendix-paraphrase} for example paraphrased prompts.
\subsubsection{New Travel Plan Constraints}
\label{sec:gen-constraint}
Since the travel planning problem involves various constraints of different types, our example instruction steps may not be comprehensive enough to cover all possible constraints. Here we examine our framework's robustness by testing whether it could generalize to the constraint types not shown in prompt examples in a zero-shot manner. As shown in Table~\ref{constraint}, UnsatChristmas has different constraints than TravelPlanner. We show that by adding several lines of constraint description in the JSON-Step prompt, LLM could generate steps for new constraints without the need to add new examples. Figure~\ref{fig:generalization} shows how our framework encodes unseen constraints in UnsatChristmas. Please see Appendix~\ref{sec:appendix-gen} for the added constraint description and see Appendix~\ref{sec:gen-ex} for full generated steps. 
\subsubsection{New Multi-Constraint Planning Tasks}
\label{sec:gen-task}
To show the capability of our framework to generalize to other domains, we conduct experiments in four new tasks: \textbf{Block Picking}, \textbf{Task Allocation}, \textbf{Travelling Salesman Problem (TSP)}, \textbf{Warehouse}. For each task, we create 25 different scenarios. See Appendix~\ref{sec:appendix-newtasks} for detailed descriptions. For both Query-Step and Step-Code generation, we include one example from travel planning and a few lines of new problem description (Appendix~\ref{sec:appendix-newtaskprompt}) in the prompt to test the zero-shot generalization capability. We implement the TwoStage tool-use framework as baselines. We use GPT-4o~\cite{gpt-4o} as the LLM to account for long code generation. We use both delivery rate and optimal rate as the evaluation metrics. Results in Table~\ref{newtasks} show that LLMs themselves fail to directly solve multi-constraint problems with large solution spaces, but our framework, with its knowledge of encoding and solving travel planning problems as SMT problems, could be adapted to other multi-constraint problems in a zero-shot manner with good optimal rate: 92\%, 92\%, 100\%, and 72\% respectively. Please refer to Appendix~\ref{sec:appendix-newtaskfail} for failure cases analysis. 
\subsection{Interactive Plan Repair Evaluation}
We examine our framework's interactive plan repair capability on both the modified queries from TravelPlanner and UnsatChristmas.\\
\textbf{Evaluation Metric}\quad We evaluate success rate: if our framework successfully modifies the query and delivers a feasible plan within 10 iterations.  \\
\textbf{Implementation Details}\quad 
Queries of UnsatChristmas have four unsatisfactory modes: 1) not enough budget, 2) no required non-stop flight, 3) no required airline, 4) no required attraction category.
We test our framework with simulated users with different preferences. One mimic user agrees to all suggestions proposed by LLM, and five mimic users have hard constraints for budget, non-stop flight, airline, attraction category, and destination cities, respectively. They refuse suggestions that change their hard constraint and provide feedback indicating they will not change this information.\\
Modified queries of TravelPlanner have three unsatisfactory modes: 1) not enough budget, 2) no required transportation method, and 3) no required house type. Mimic users have hard constraints for budget, destination cities, transportation methods, and house type. \\
\textbf{Ablation Studies}\quad
Key components in our framework include 1) LLM receives unsatisfiable reasons from the solver; 2) LLM collects information based on the reasons, analyzes, and offers suggestions; 3) LLM receives human preferences regarding offered suggestions and modifies codes; 4) SMT solver gives satisfiability verification. We perform ablation studies to examine these key components. We compare: 
1) \textbf{No Reason}: asking LLM to resolve unsatisfiable queries without providing unsatisfiable reasons; 2) \textbf{No Feedback}: asking the human to only provide binary \textit{``agree''} or \textit{``disagree''} feedback without explaining why; 3) \textbf{No Solver}: removing the SMT solver. The LLM directly gives a list of suggestions in one iteration because no solver is provided to verify the updated query; 4) \textbf{Ours}: our approach (Section~\ref{sec:planrepair}); 5) \textbf{Ours-20}: our approach with maximum 20 iterations.\\
\textbf{Results and Analysis}\quad Table~\ref{unsat_result_1} and~\ref{unsat_result_2} show the interactive plan repair performance. Our framework successfully addresses an average of 78.6\% and 85.0\% diverse human preferences across all types of mimic humans. \textbf{Ours-20} raises the success rate to 81.6\% and 91.7\%, showing the potential of increasing iteration limits to achieve better results. We include figures of iterations versus performance in Appendix~\ref{sec:appendix-iteration}. For queries from both datasets, \textbf{Ours} significantly outperforms \textbf{No Solver} by an average of 53.4\% and 60.0\% across all mimic humans. This suggests that LLM's capability to utilize the SMT solver to verify the modified query largely benefits the interactive plan repair process. \textbf{Ours} also outperforms \textbf{No Reason} by an average of 14.9\% and 5.0\% and outperforms \textbf{No Feedback} by an average of 7.8\% and 14.1\%. These results validate the effectiveness of our key components.

\section{Conclusion}
In this work, we propose a framework that enables LLMs to utilize an SMT solver to formally formulate and solve complex realistic planning problems as constrained satisfiability problems. Our framework generalizes to natural language query inputs, almost guarantees to deliver plans for satisfiable queries with a pass rate of 93.9\%, and provides personalized suggestions to modify unsatisfiable queries. We prove our framework can handle diverse paraphrased prompts. We also show that our framework can generalize to unseen constraint types and new domains in a zero-shot manner.  

\section{Limitation}
The limitations and potential risks of the work are as follows:\\
\textbf{Prompt Designing}\quad We need a careful design of instruction steps and corresponding codes to encode the problem. It is time-consuming to formulate the problem from scratch. However, as discussed in Section~\ref{sec:related-prompt} \textbf{LLM Prompt Design}, most of the existing works that could achieve strong performance on complex planning problems rely on different forms of task-specific efforts. In addition, the potential of our framework to generalize to the unseen constraints and unseen tasks eases the future efforts needed to incorporate more constraints into the framework and to solve more different multi-constraint problems. We also show our framework compatibility to handle diverse paraphrased prompts. Moreover, as the designers of the framework, we design offline prompts to enable full model autonomy for end users. Thus, after these prompts are designed, the effort needed for any end user to utilize our framework is a simple natural language query. With our framework, end users can utilize powerful solvers to solve their problems without having any knowledge about the solvers.\\
\textbf{Solver Runtime}\quad The runtime of SMT may become slower as the problem complexity increases. For the TravelPlanner dataset, we set the maximum runtime of the SMT solver to 30 minutes. But we want to emphasize that only 1.3\% of the 1180 queries fail to find a plan because their runtime exceeds the 30-minute limit we set. Except for that, for 179 out of 180 solved queries in the validation set, the solver on average takes 38.39 seconds to solve one query. We include a detailed runtime and cost analysis in Appendix~\ref{sec:appendix-time}. For more massive databases with more destination city choices, various constraint types, and queries that only have a few feasible plans, our framework could take a long runtime to find the plan. To relieve this limitation, a potential way is to introduce some heuristics and prioritize a portion of the the choices to be verified first. In addition, since SMT solvers tend to explore a large logical search space, other solvers could be faster for certain types of problems. For example, if the problem has purely linear constraints and optimization goals, mixed-integer linear program (MILP) solvers are likely to be faster than SMT solvers. We believe there is great potential to adapt the framework to use alternative solvers for runtime advantages if preferred. For example, encoding with the MILP solver is similar to the SMT solver in that they both follow the [variable initialization - adding constraints - (possibly) calculating and optimizing objectives]. Following our framework, one simple solution is introducing a new component in the framework that, provided with some SMT->MILP translation codes, asks LLMs to base on previously generated steps and SMT codes to write the MILP codes. We are happy to explore more possibilities in the future.\\
\textbf{Risky Data}\quad Since all information sources of our framework is from the database we use, it currently does not have the capability to distinguish unsafe or incorrect information. One potential risk of our framework is that it could generate risky plans based on unsafe information from the database.

\section*{Acknowledgments}
This work was supported by ONR under Award N00014-22-1-2478 and MIT-IBM Watson AI Lab. However, this article solely reflects the opinions and conclusions of its authors.

\bibliography{custom}

\onecolumn
\newpage
\addtocontents{toc}{\protect\setcounter{tocdepth}{2}}
\renewcommand{\contentsname}{Large Language Models Can Solve Real-World Planning Rigorously with Formal Verification Tools}
\tableofcontents 

\appendix
\newpage
\section{Description of constraints}
\begin{table*}[!ht]
  \centering
  \begin{tabularx}{\textwidth}{l|X}
    \hline
    \textbf{Constraint}           & \textbf{Description} \\
    \hline
    Destination cities      & Destination cities should not be repeated \\
    \hline
    Transportation dates    &  First transportation happens at first day, last transportation happens at last day, and others happens in between non-repeatedly \\
    \hline
    Transportation methods &  \textcolor{teal}
    {Every transportation uses flight, self-driving, or taxi\hfill \break Self-driving is not valid during the trip if taxi or flight is used\hfill \break No flight if \textit{``no flight''} is mentioned, and no self-driving if \textit{``no self-driving''} is mentioned} \\
    \hline
    Flight & \textcolor{teal}{No flight if flights unavailable between two cities on certain dates}\hfill \break \textcolor{brown}{All taken flights are non-stop if \textit{``non-stop''} is mentioned\hfill \break All taken flights' airlines are within the required airlines list}\\
    \hline
    Driving & \textcolor{teal}{No driving if driving routes unavailable between two cities}\\
    \hline
    Restaurant &\textcolor{teal}{Restaurant choices should not be repeated\hfill \break Restaurant for day must be located within that day’s city\hfill \break All specified cuisine types must be visited}\\
    \hline
    Attraction & Attraction choices should not be repeated\hfill \break Attraction for day must be located within that day’s city\hfill \break \textcolor{brown}{All specified attraction types must be visited}\\
    \hline
    Accommodation & \textcolor{teal}{Accommodation for day must be located within that day’s city\hfill \break All specified accommodations must satisfy specified Room Rule\hfill \break All specified accommodations must satisfy specified Room Type\hfill \break The number of consecutive days spent in an accommodation must meet the accommodation's minimum number of nights’ stay.}\\
    \hline
    Budget & Total spend is within specified budget\\
    \hline
\end{tabularx}
  \caption{\label{constraint}
    Descriptions of constraints for two datasets. Constraints in \textcolor{teal}{teal} are the constraints only in TravelPlanner. Constraints in \textcolor{brown}{brown} are the constraints only in our dataset. Constraints in black are common constraints.
  }
\end{table*}
\clearpage
\section{Runtime and cost analysis}
\label{sec:appendix-time}
Since TravelPlanner's database has 65 states, 312 cities, 3827361 flights, 17603 driving information, 5303 attractions, 9552 restaurants, and 5064 accommodations, the solution space is extremely large considering the combinatorial choices. In addition, a few queries are challenging in that they have few feasible plans. We limit SMT solver's maximum runtime for each query to 30 minutes. However, we show below that solver is not time consuming for most of the cases.\\
In this section, we include the detail runtime and cost analysis of both satisfiable plan solving and interactive plan repair of our framework.
\subsection{Satisfiable Plan Solving}
For the satisfiable plan solving part, we recorded the runtime and cost for 180 queries in TravelPlanner’s validation set. Over the 180 queries, the average cost is \$0.74 per query using GPT-4. Over the 179 queries with delivered plans, the average time spent for different stages in our framework are shown in Table~\ref{runtime1}. The average total time spent for all stages is 245.66 seconds (4.09 minutes) per query. The Step-Code generation contains multiple LLM calls for various types of constraints, thus takes most of the time.
\begin{table*}[!ht]
\begin{small}
\begin{tabularx}{\textwidth}{*{4}{Y}|Y}
\toprule
 LLM NL-JSON & LLM JSON-Step & LLM Step-Code & SMT Solver & Total \\
\midrule
5.45 & 35.16 & 166.66 & 38.39 & 245.66 \\
\bottomrule
\end{tabularx}
\caption{\label{runtime1}
    Runtime (seconds) of each stage of our framework for satisfiable plan solving.
}
\end{small}
\end{table*}
\\
Out of the 180 queries, there is one query with no delivered plan since its runtime exceeds 30 minutes. For queries with heavy computational costs, introducing some heuristics that prioritize a portion of all possible solutions could help to reduce the computational overhead of SMT solvers. Our framework introduces a simple heuristic: for queries that ask to visit multiple cities in a state, we will prioritize the cities with available transportations between the origin. This heuristic helps to reduce the runtime, especially for a big state with ~20 cities. In addition to this simple heuristic, some other heuristics may help, which we plan to explore more in the future: pre-calculate estimated money spent and prioritize the cheaper solutions, prioritize the cities with a larger number of transportation methods/ restaurants/ accommodations, etc.
\subsection{Interactive Plan Repair}
For the interactive plan solving part, we recorded the runtime and cost for queries in UnsatChristmas for mimic-human with hard budget constraints. Over the 23 (out of 39) successful queries, the average cost is \$0.65 per iteration using GPT-4. The average time spent for different stages in our framework are shown in Table~\ref{runtime2}. The average total time spent for both stages is 33.68 seconds per iteration. Note that for mimic-human with hard budget constraints, the average number of iterations that successfully modify the queries is 2.22 per query. 
\begin{table*}[!ht]
\begin{tabularx}{\textwidth}{*{2}{Y}|Y}
\toprule
 LLM interactive suggestion  & Code Update & Total \\
\midrule
10.35 & 23.33 & 33.68 \\
\bottomrule
\end{tabularx}
\caption{\label{runtime2}
    Runtime (seconds per iteration) of each stage of our framework for interactive plan repair.
}
\end{table*}

\newpage
\section{Example input queries and output plans}
\label{sec:appendix-ex}
 In an query, the user can specify 1) length of travel (3, 5, or 7 days), 2) the destination city or state (for 5/7 days travel, the destination cities would be 2/3 cities from a state), 3) travel dates, 4) budget, 5) preferences regarding transportation methods, 6) preferences regarding restaurant cuisine types, 7) preferences regarding accommodation type and rules. \\
 We list an example input query and the corresponding output plans.\\
\begin{boxC}
\textbf{Input query:}\\
Can you create a 5-day travel itinerary for a group of 3, departing from Atlanta and visiting 2 cities in Minnesota from March 3rd to March 7th, 2022? We have a budget of \$7,900. We require accommodations that allow parties and should ideally be entire rooms. Although we don't plan to self-drive, we would like the flexibility to host parties.\\
\textbf{Corresponding output plan:} \\
\{\\
    "days": 1, \\
    "current\_city": "from Atlanta to Minneapolis(Minnesota)", \\
    "transportation": "Taxi, from Atlanta to Minneapolis(Minnesota), duration: 16 hours 26 mins, distance: 1,821 km, cost: 1821", \\
    "breakfast": "-", \\
    "attraction": "Minneapolis Sculpture Garden, Minneapolis(Minnesota)", \\
    "lunch": "-", \\
    "dinner": "Bawa Chicken, Minneapolis(Minnesota)", \\
    "accommodation": "Light-filled Apartment in great area, close 2 all!, Minneapolis(Minnesota)"\\
\}, \\
\{\\
    "days": 2, \\
    "current\_city": "from Minneapolis(Minnesota) to St. Cloud(Minnesota)", \\
    "transportation": "Taxi, from Minneapolis(Minnesota) to St. Cloud(Minnesota), duration: 1 hour 6 mins, distance: 105 km, cost: 105", \\
    "breakfast": "Friends Restaurant, St. Cloud(Minnesota)", \\
    "attraction": "Blacklight Adventures, St. Cloud(Minnesota)", \\
    "lunch": "Madras Cafe, St. Cloud(Minnesota)",\\ 
    "dinner": "Aggarwal's Sweets Paradise, St. Cloud(Minnesota)", \\
    "accommodation": "The Gramercy East, St. Cloud(Minnesota)"\\
\}, \\
\{\\
    "days": 3, \\
    "current\_city": "St. Cloud(Minnesota)", \\
    "transportation": "-", \\
    "breakfast": "Mocha, St. Cloud(Minnesota)",\\ 
    "attraction": "AirMaxx Trampoline Park \& Warrior Course, St. Cloud(Minnesota)", \\
    "lunch": "House of Commons, St. Cloud(Minnesota)", \\
    "dinner": "Cakewalk - Park Plaza, St. Cloud(Minnesota)", \\
    "accommodation": "The Gramercy East, St. Cloud(Minnesota)"\\
\}, \\
\{\\
    "days": 4, \\
    "current\_city": "St. Cloud(Minnesota)", \\
    "transportation": "-", \\
    "breakfast": "Pa Pa Ya, St. Cloud(Minnesota)",\\ 
    "attraction": "Beaver Island Trail South, St. Cloud(Minnesota)", \\
    "lunch": "Grecian Key Restaurant, St. Cloud(Minnesota)", \\
    "dinner": "Food Care, St. Cloud(Minnesota)", \\
    "accommodation": "The Gramercy East, St. Cloud(Minnesota)"\\
\}, \\
\{\\
    "days": 5, \\
    "current\_city": "from St. Cloud(Minnesota) to Atlanta", \\
    "transportation": "Taxi, from St. Cloud(Minnesota) to Atlanta, duration: 17 hours 19 mins, distance: 1,919 km, cost: 1919", \\
    "breakfast": "Annapurna Sweets, St. Cloud(Minnesota)", \\
    "attraction": "-", \\
    "lunch": "Republic of Chicken, St. Cloud(Minnesota)", \\
    "dinner": "-", \\
    "accommodation": "-"\\
\}
\end{boxC}

\newpage
\section{Satisfiable Plan Solving Evaluation Details}

\subsection{Result of Ours+JSON on TravelPlanner}
\label{sec:appendixcours+json}
Here we include the Ours+JSON framework and satisfiable Plan Solving result of Ours+JSON on dataset TravelPlanner. We test the same LLMs (GPT-4, Claude-3-Opus, and Mistral-Large) on the validation set and test GPT-4 on the test set of TravelPlanner. Our framework achieves final pass rates of 98.9\%, 98.3\%, and 84.4\% respectively on validation set, and 97.0\% for GPT-4 on test set. From the result, the performance is further improved by 5.6\%, 5.0\%, and 17.7\% for GPT-4, Claude-3, and Mistral on validation set, and is further improved by 6.8\% for GPT-4 on test set. Since translating the JSON allows the LLMs to first extract and summarize the key information from the natural langauge query into a fix-formatted clear representation, it helps to understand and represent the problem better.
\begin{table*}[!ht]
\caption{Result comparison of Ours and Ours+JSON with three LLMs on TravelPlanner}
\label{json}
\begin{center}
\vspace{-8pt}
\begin{small}
\begin{tabularx}{\textwidth}{lYY|YY}
\toprule
 & 
Valid (Ours) & Valid (Ours+JSON) & Test (Ours) & Test (Ours+JSON) \\
\midrule
GPT-4
 & 93.3 & 98.9  & 90.2 & 97.0 \\
Claude-3
 & 93.3 & 98.3 & 93.9 & N/A  \\
Mistral-Large
 & 66.7 & 84.4 & 67.8 &  N/A \\

\bottomrule
\end{tabularx}
\end{small}
\end{center}
\vspace{-6pt}
\end{table*}

\begin{figure*}[!ht]
\center\includegraphics[width=0.9\linewidth]{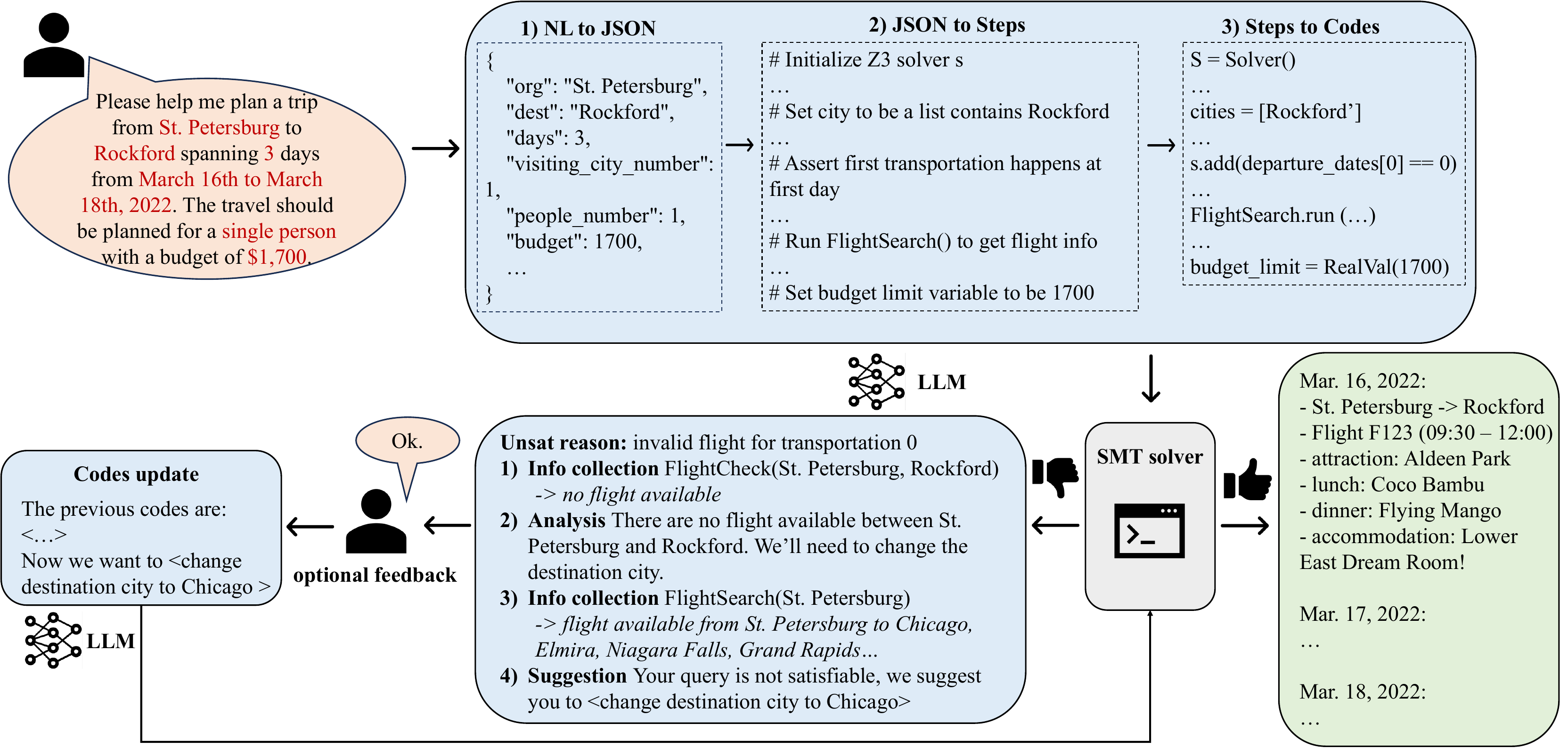} 
  \caption {An overview of the framework with JSON.}
  \label{fig:Framework_json}
  \vspace{-12pt}
\end{figure*}

\subsection{Description of other evaluation metrics}
\label{sec:appendixctype}
Delivery Rate measures whether a final plan is generated within a limited time. \\
Commonsense constraints defined in~\cite{xie2024travelplanner} include: all information in the plan is within closed sandbox, the plan is complete without any left out part, all activities should be conducted in current city, travel route is reasonable, restaurant and attractions should not be repeated, transportation is reasonable (no self-driving if taxi or flight is taken during the travel), the nunmber of consecutive days spent in a specific accommodation must meet its required minimum number of nights' stay. \\
Hard constraints include: the total spend of the trip is within budget, the specified room rule does not exist ("No parties”, “No smoking”, “No children under 10”, “No pets”, and “No visitors”), the specified room type exists (“Entire Room”, “Private Room”, “Shared Room”, and “No Shared Room”), the specified cuisine types are fulfilled during the trip (“Chinese”, “American”, “Italian”, “Mexican”, “Indian”, “Mediterranean”, and “French”), the specified transportation method is satisfied (“No flight” and “No self-driving”.).\\
For Commonsense Constraint Pass Rate and Hard Constraint Pass Rate, two evaluation modes, micro and macro, are used to test the agent's capability to follow single constraint and follow constraints holistically. Micro calculates the ratio of passed constraints to the total number of constraints, while Macro calculates the ratio of plans that pass all commonsense or hard constraints among all tested plans.

\newpage
\section{Interactive Plan Repair: Iteration versus Performance}
\label{sec:appendix-iteration}
\begin{figure*}[!ht]
  \includegraphics[width=\linewidth]{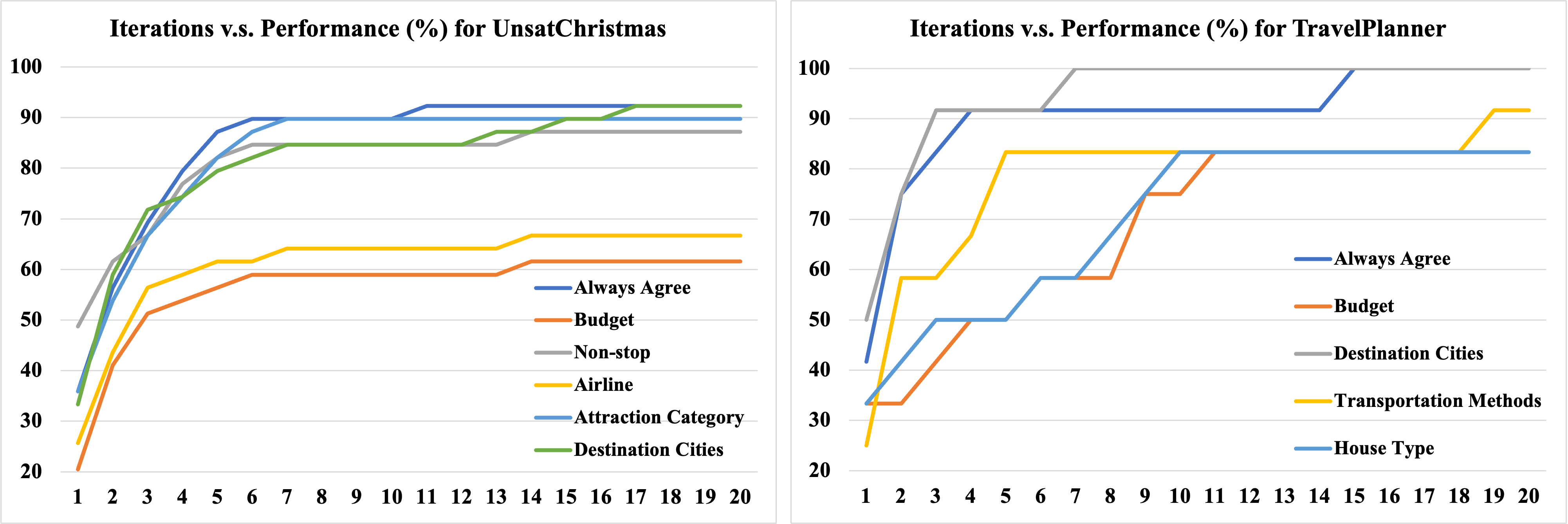} 
  \caption {Performance (success rate \%) of interactive plan repair over different numbers of iterations for two datasets}
  \label{fig:iterations}
\end{figure*}

Figure~\ref{fig:iterations} shows the performance (success rate \%) of interactive plan repair over different numbers of iterations for both datasets. \\
For the 39 queries in UnsatChristmas, 63.7\% of the queries could be successfully modified to be satisfiable within 3 iterations, 74.8\% within 5 iterations, 78.6\% within 10 iterations, and 81.6\% within 20 iterations. The performance increases quickly during the first 5 iterations, and the framework solves a limited number of more difficult queries with more iterations. \\
Similarly, for the 12 modified queries in TravelPlanner, 65.0\% of the queries could be successfully modified to be satisfiable within 3 iterations, 73.3\% within 5 iterations, 85.0\% within 10 iterations, and 91.7\% within 20 iterations. \\
The results suggest that we do not need extensive iterations to fully capture a major portion of the human queries.

\newpage
\section{New Multi-constraint Tasks details}
\subsection{Task Setup}
\begin{figure*}[!ht]
\center\includegraphics[width=0.9\linewidth]{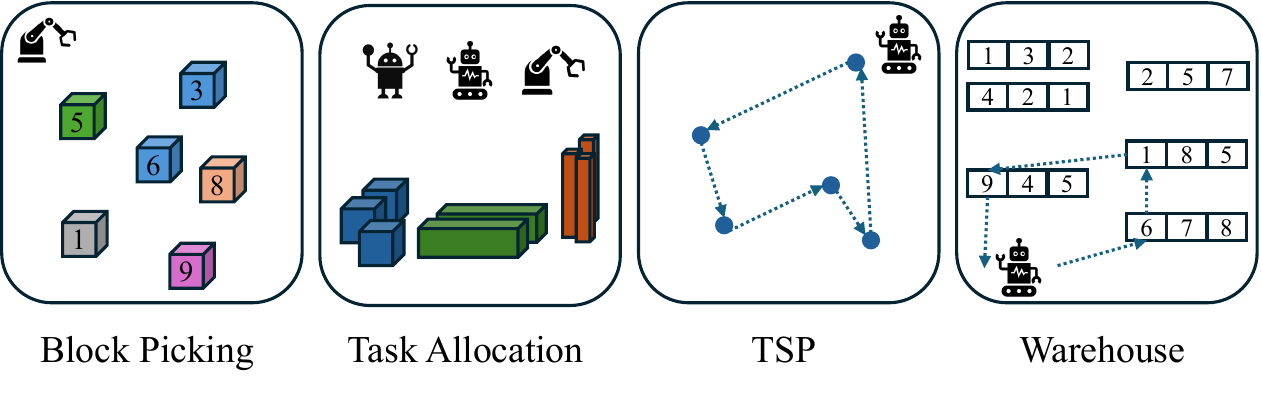} 
  \caption {Four new multi-constraint tasks.}
  \label{fig:gentasks}
  \vspace{-12pt}
\end{figure*}
\label{sec:appendix-newtasks}
\subsubsection{Block Picking}
There are blocks of different colors and scores in the scene. The goal is to select required number of unique blocks with required color, while maximizing the score. All possible block colors are red, yellow, black, pink, and blue. For 25 different scenarios, we set the total number of blocks to be a random number between 50 to 200, each with a random score between 1 to 20 and a random color. For the query, we will randomly choose 1 to 3 colors from all colors to be the required color, and 10 to be the required number of blocks to pick.

\subsubsection{Task Allocation}
Given a list of three tasks A, B and C, and three heterogeneous robots that are skilled at different tasks, the goal is to find the way to assign different tasks to different robots and finish the tasks with minimized finish time. The three robots could work in parallel, but the finish time counts the time when the last robot stops working. For 25 different scenarios, we set number of task A, B and C to be random numbers between 10 to 100. For each robot, we set its work time to finish each task to be a random number between 10 to 100. 
\subsubsection{TSP}
Given a list of ten cities, the goal is to visit each city exactly once with minimized distance travelled. For 25 different scenarios, we set the coordinates of each city to be a random tuple between 0 and 1.

\subsubsection{Warehouse}
The robot has a task list of length N that needs to be finished one by one. In the warehouse, there are 50 stations, where the robots can visit stations to finish different tasks. 
The robot starts at station 0, travel n stations to finish n tasks, and then travel back to station 0.
The robot needs to finish n tasks while minimizing the total distance travelled. 
For 25 different scenarios, we set the total task number to be 50, and the total station number to be 50, and each station can be used to accomplish 3 random tasks. We set the number of tasks the robot need to finish to be a random number from 3 to 10, and the tasks to be random numbers within 50. We set the coordinates of each station to be a random tuple between 0 and 1.

\subsection{Added task description in prompt}
\label{sec:appendix-newtaskprompt}
\begin{boxC}
\textbf{Block Picking}\\
Now, you are given a JSON constraint of a block stacking problem. \\
There are blocks of different colors and scores in the scene. You need to select "block\_number" non-repeat blocks with color in "color" list, while maximizing the score.\\
You have access to a BlockSearch API.\\ 
BlockSearch.run(color:list) gives 1.all possible block ids of color in "color" list and 2.corresponding score info. You should assert chosen blocks index does not exceed block id list length.\\ 
BlockSearch.get\_info(score\_info, block\_index) gives the score of certain block. \\\\
\textbf{Task Allocation}\\
Now, you are given a JSON constraint of a task allocation problem.\\
Given a list of tasks and three heterogeneous robots that are skilled at different tasks, the goal is to find the way to assign different number of tasks to different robots and finish the tasks with minimized finish time. The three robots could work in parallel, but the finish time counts the time when the last robot stops working. \\
You have access to a TimeSearch() API. \\
TimeSearch.run() searches robots' accomplishing time info.\\
TimeSearch.get\_info(time\_info, robot: str, task: str) gives the accomplishing time of certain block for certain task. An example of robot and task string is: 'robotA', 'taskA'.\\
Note that for each robot R and each task T, the number of task T robot R is allocated needs to be non-negative and within the total number of T.\\
You have access to a Max(variable\_list) function that outputs the max of a list of variables.\\\\
\textbf{TSP}\\
Now, you are given a JSON constraint of a travelling salesman problem (TSP) problem. \\
Given 10 cities, you need to non-repeatly visit each city exactly once with minimized distance travelled. \\
You have access to a DistanceSearch() API. \\
DistanceSearch.run() takes no argument and gives the distance info between cities, and DistanceSerarch.get\_info(distance\_info, city\_1, city\_2) gives the distance(a real number) between two cities. \\
You should explicitly assert city index does not exceed total number of cities.\\
You do not need to count the distance to go back to the first city, so the total number of distances you need to consider is 9. \\\\
\textbf{Warehouse}\\
Now, you are given a JSON constraint of a warehouse robot routing problem. \\
The robot are given a task list "task\_id" of length n and needs to finish them one by one. In the warehouse, there are "total\_station\_number" stations, where the robots can visit stations to finish different tasks. \\
The robot needs to finish n tasks while minimizing the total distance travelled. When calculating total travel distance, make sure to include 1.the distance to travel from origin (0) to first station; 2.distance between n stations (so this is n-1 values); 3.distance to travel from last station back to origin(0).\\
You have access to a StationSearch() API. \\
StationSearch.run\_task(tasks) takes a list of tasks that the robot needs to accomplish, and gives a list of stations\_id\_list. For each task, you should assert robot choose one station id from corresponding stations\_list, which is all possible stations.\\
StationSearch.run\_distance() takes no argument and gives the distance info between stations, and StationSearch.get\_info(distance\_info, station\_1\_id, station\_2\_id) takes gives the distance (a real number) between two stations. \\
\end{boxC}
\subsection{Failure case analysis}
\label{sec:appendix-newtaskfail}
For Block Picking, Task Allocation, and Warehouse, there are failure cases. \\\\
\textbf{Block Picking} For Block Picking, LLM fails to give the optimal plan for 2 out of 25 delivered plans. The failure reasons are same for these two plans. In the block picking task, all picked blocks need to be distinct. Thus, in the codes LLM writes, it needs to explicitly check all block indexes it chooses are different. In these two plans, the LLM fails to take this into account, thus repeatedly choose same blocks with high scores to maximize the score.\\\\
\textbf{Task Allocation} For Task Allocation, LLM fails to deliver the plan for 2 out of 25 scenarios. The failure reasons are same. The LLMs are provided a Max(variable\_list) function, which takes in a list of variables and return the max. However, in the codes written by LLM, they fail to input a list of variables, but input the variables themselves one by one. This gives runtime errors thus fails to deliver the plan.\\\\
\textbf{Warehouse} For Warehouse, LLM fails to deliver the plan for 4 out of 25 scenarios, and fails to give the optimal plan for 3 out of the 21 delivered plan.\\
Note that Warehouse is a more challenging task in that it needs to select stations to visit while calculate the minimum travel distance between them at the same time. Thus, the codes are more complex to write. \\
The delivery failure reasons for the 4 scenarios are same. Since the task requires the robot to travel from origin, visit n stations, and back to origin at the end. LLM could choose to set n station variables to represent n stations needs to visit or set n+2 station variables and make the first and last one to equal to 0. However, in the failure cases, the LLM set n variables to represent n stations, but at the same time assert first and last one to equal to 0. This brings conflicts because the it could assert one variable to equal to two values at the same time.\\
The non-optimal reasons for the 3 delivered plan are same: since the StationSearch.run\_task(tasks) outputs a station\_id\_list, the contents are IDs of stations. LLM needs to create station variables and assert it to equal to \textit{value} from the ID lists. However, in the codes, LLM assert the station variables to equal to \textit{index} from zero to length of station\_id\_list. This makes it to calculate incorrect distance thus outputting non-optimal solutions. 

\newpage
\section{Prompts}
\label{sec:appendix-prompt}
\subsection{Prompts for Satisfiable Plan Solving}
\label{sec:appendix-sat}
\subsubsection{Query-Step prompt}
\label{sec:appendix-sat1}
The instruction prompt for natural language to steps translation prompt is provided as follows: 
\begin{boxC}
You are given a natural language query that contains important constraints to satisfy for a travel plan problem. These constraints could include departure city, destination state or city, total travel days and dates, and some spcecial requirements regarding accommodation, restaurant, and transportation. \\
Your job is to give a detailed step by step instuction to encode this constraint as code. \\
Here are some example steps for different constraint:\\
-----EXAMPLE 1-----\\
Natural Language query:\\
Can you create a 5-day travel itinerary for a group of 3, departing from Atlanta and visiting 2 cities in Minnesota from March 3rd to March 7th, 2022? We have a budget of \$7,900. We require accommodations that allow parties and should ideally be entire rooms. Although we don't plan to self-drive, we would like the flexibility to host parties.\\
Steps:\\
\# Destination cities \# \\
\# Run CitySearch to get all possible destination cities in Minnesota State for dates ["2022-03-16", "2022-03-17", "2022-03-18"] from origin 'Atlanta', remove origin 'Atlanta' if it is in list\\
\# Loop through cities for 2 destination cities\\
\# Initialize Z3 solver s\\
\# Set 'city' variable to be indexes of 2 destination cities\\
\# If city\_0\_index and city\_1\_index are not same, assert 2 'city' variables equal to city index\\

\# Departure dates \# \\
\# Set 'departure\_dates' variables for 3 transportations between cities\\
\# Assert first transportation happens at first day (day 0), last transportation happens at last day (day 4), and second transportation could happen at any day in between\\

\# Transportation methods \# \\
\# Set transportation method (flight, self-driving, taxi) variable for 3 transportations between cities\\
\# Assert only one of flight, self-driving, or taxi is used for 3 transportations between cities, self-driving is not valid if taxi or flight is used for any transportation\\
\# Assert all 3 transportations between cities are not self-driving\\

\# Flight information \# \\
\# Run FlightSearch to get flight info for Atlanta as origin, list of cities, city\_0 and city\_1, and dates ["2022-03-16", "2022-03-17", "2022-03-18"]\\
\# Get specific flight price info with Atlanta as origin and final destination, specific city variable, and departure date for 3 transportations\\
\# Set 'flight\_index' variable for 3 transportations\\
\# Assert 3 'flight\_index' variables are within valid range if taking flight, assert flight index to be -1 if not taking flight\\
\# Calculate flight price for 3 people for 3 transportations based on flight index variable\\
\# Get specific flight arrival time info with Atlanta as origin and final destination, specific city, and departure date for 3 transportations\\
\# Calculate flight arrival time for 3 transportations based on flight index variable\\

\# Driving information \# \\
\# Run DistanceSearch to get driving info for Atlanta as origin and city\_0 and city\_1\\
\# Get specific driving distance info with Atlanta as origin and final destination, specific city, and departure date for 3 transportations\\
\# Assert driving info is not empty if driving\\
\# Calculate self-driving and taxi price for 3 people and 3 transportations based on driving distance\\
\# Get driving arrival time with Atlanta as origin and final destination, specific city, and departure date for 3 transportations\\

\# Restaurant information \# \\
\# Get arrivals and city list for each day based on 3 transportations, 5 total travel day, and departure dates variables\\
\# Run RestaurantSearch to get restaurant price info and cuisine info for city\_0 and city\_1\\
\# Set 'restaurant\_in\_which\_city' variables for 15 (3 meals per day, 5 days) meals\\
\# For each 'restaurant\_in\_which\_city' variable, assert it to be either current city or next city based on transportation arrivals time\\
\# Set 'restaurant\_index' variables for 15 (3 meals per day, 5 days) meals\\
\# For each 'restaurant\_index', get specific price info based on 'restaurant\_in\_which\_city' variable, assert index are within valid range, assert restaurants in same city are not repeated, and calculate restaurant price for 3 people\\
\# Calculate restaurant price based on restaurant index\\

\# Attraction information \# \\
\# Run AttractionSearch to get attraction info for city\_0 and city\_1\\
\# Set 'attraction\_in\_which\_city' variables for 5 (1 per day) attractions\\
\# For each 'attraction\_in\_which\_city' variable, assert it to be either current city or next city based on transportation arrivals time\\
\# Set 'attraction\_index' variables for 5 (1 per day) attractions\\
\# For each 'attraction\_index', get specific length info based on attraction in which city variable, assert index are within valid range, and attrations in same city are not repeated\\

\# Accommodation information \# \\
\# Run AccommodationSearch to get accommodation info and accommodation constraints for city\_0 and city\_1\\
\# Set 'accommodation\_index' variables for 2 (1 per city) accommodations\\
\# For each 'accommodation\_index', get specific price info based on accommodation in which city variable, assert 'accommodation\_index' variable are within valid range, calculate number of room need for 3 people and accommodation price\\
\# For each city, get accommodation minimum night info and assert it to be less than the days stay in this city\\
\# For each 'accommodation\_index', get specific room type and house rules info, assert 'Entire home/apt' exist for all accommodations, assert 'No parties' does not exist for all accommodations\\

\# Budget \# \\
\# Set budget limit variable to be 7900\\
\# Add 3 transportation price to spent, according to whether transportation method is flight, self-driving, or taxi\\
\# Add restaurant price to spent\\
\# Add accommodation price to spent\\
\# Assert current spent is within budget\\

-----EXAMPLE 2-----\\
Natural Language query:\\
Can you help with generating a 7-day travel plan for a party of 5? We're setting off from Indianapolis and planning to explore 3 cities in Colorado from March 11th to March 17th, 2022. We have a budget of \$15,100 for this trip. We'll be bringing our pets, so pet-friendly accommodations are a must. It's important for us to stay in accommodations that permit children under the age of 10. We're also hoping to find places that offer Mexican, Italian, Mediterranean, and Indian cuisines. No shared rooms for accommodations would be ideal. We do not have preferences for transportation.\\
Steps:\\
\# Destination cities \# \\
\# Run CitySearch to get all possible destination cities in Colorado State for dates ['2022-03-11', '2022-03-12', '2022-03-13', '2022-03-14', '2022-03-15', '2022-03-16', '2022-03-17'] from origin 'Indianapolis', remove origin 'Indianapolis' if it is in list\\
\# Loop through cities for 3 destination cities\\
\# Initialize Z3 solver s\\
\# Set 'city' variable to be indexes of 3 destination cities\\
\# If city\_0\_index, city\_1\_index, city\_2\_index are not same, assert 3 'city' variables equal to city index\\

\# Departure dates \# \\
\# Set 'departure\_dates' variables for 4 transportations between cities\\
\# Assert first transportation happens at first day (day 0), last transportation happens at last day (day 6), second and third transportation happen in between but not at the same day\\

\# Transportation methods \# \\
\# Set transportation method (flight, self-driving, taxi) variable for 4 transportations between cities\\
\# Assert only one of flight, self-driving, or taxi is used for 4 transportations between cities, self-driving is not valid if taxi or flight is used for any transportation\\

\# Flight information \# \\
\# Run FlightSearch to get flight info for Indianapolis as origin, list of cities, city\_0, city\_1 and city\_2, and dates ['2022-03-11', '2022-03-12', '2022-03-13', '2022-03-14', '2022-03-15', '2022-03-16', '2022-03-17']\\
\# Get specific flight price info with Indianapolis as origin and final destination, specific city variable, and departure date for 4 transportations\\
\# Set 'flight\_index' variable for 4 transportations\\
\# Assert 4 'flight\_index' variables are within valid range if taking flight, assert flight index to be -1 if not taking flight\\
\# Calculate flight price for 5 people for 4 transportations based on flight index variable\\
\# Get specific flight arrival time info with Indianapolis as origin and final destination, specific city, and departure date for 4 transportations\\
\# Calculate flight arrival time for 4 transportations based on flight index variable\\

\# Driving information \# \\
\# Run DistanceSearch to get driving info for Indianapolis as origin and city\_0, city\_1 and city\_2\\
\# Get specific driving distance info with Indianapolis as origin and final destination, specific city, and departure date for 4 transportations\\
\# Assert driving info is not empty if driving\\
\# Calculate self-driving and taxi price for 5 people and 4 transportations based on driving distance\\
\# Get driving arrival time with Indianapolis as origin and final destination, specific city, and departure date for 4 transportations\\

\# Restaurant information \# \\
\# Get arrivals and city list for each day based on 4 transportations, 7 total travel day, and departure dates variables\\
\# Run RestaurantSearch to get restaurant price info and cuisine info for city\_0, city\_1 and city\_2\\
\# Set 'restaurant\_in\_which\_city' variables for 21 (3 meals per day, 7 days) meals\\
\# For each 'restaurant\_in\_which\_city' variable, assert it to be either current city or next city based on transportation arrivals time\\
\# Set 'restaurant\_index' variables for 21 (3 meals per day, 7 days) meals\\
\# For each 'restaurant\_index', get specific price info based on 'restaurant\_in\_which\_city' variable, assert index are within valid range, assert restaurants in same city are not repeated, and calculate restaurant price for 5 people\\
\# Set 'cuisine\_type' variables for each cuisine type ['Mexican', 'Italian', 'Mediterranean', 'Indian'] required\\
\# For each cuisine type, iterate through all restaurant to check if it is satisfied\\

\# Attraction information \# \\
\# Run AttractionSearch to get attraction info for city\_0, city\_1 and city\_2\\
\# Set 'attraction\_in\_which\_city' variables for 7 (1 per day) attractions\\
\# For each 'attraction\_in\_which\_city' variable, assert it to be either current city or next city based on transportation arrivals time\\
\# Set 'attraction\_index' variables for 7 (1 per day) attractions\\
\# For each 'attraction\_index', get specific length info based on attraction in which city variable, assert index are within valid range, and attrations in same city are not repeated\\

\# Accommodation information \# \\
\# Run AccommodationSearch to get accommodation info and accommodation constraints for city\_0, city\_1 and city\_2\\
\# Set 'accommodation\_index' variables for 3 (1 per city) accommodations\\
\# For each 'accommodation\_index', get specific price info based on accommodation in which city variable, assert 'accommodation\_index' variable are within valid range, calculate number of room need for 5 people and accommodation price\\
\# For each city, get accommodation minimum night info and assert it to be less than the days stay in this city\\
\# For each 'accommodation\_index', get specific room type and house rules info, assert 'Shared room' does not exist for all accommodations, assert 'No pets' does not exist for all accommodations, assert 'No children under 10' does not exist for all accommodations\\

\# Budget \# \\
\# Set budget limit variable to be 15100\\
\# Add 4 transportation price to spent, according to whether transportation method is flight, self-driving, or taxi\\
\# Add restaurant price to spent\\
\# Add accommodation price to spent\\
\# Assert current spent is within budget\\

-----EXAMPLES 3-----\\
Natural Language query:\\
Please create a 3-day travel itinerary for 2 people beginning in Fort Lauderdale and ending in Milwaukee from the 8th to the 10th of March, 2022. Our travel budget is set at \$1,100. We'd love to experience both American and Chinese cuisines during our journey. We'd love to live in private rooms but we don't want to take flights. \\
Steps:\\
\# Destination cities \# \\
\# Set cities to be a list includes Milwaukee only\\
\# Loop through cities for 1 destination cities\\
\# Initialize Z3 solver s\\
\# Set 'city' variable to be indexes of 1 destination cities\\
\# Assert 'city' variable equal to city index\\

\# Departure dates \# \\
\# Set 'departure\_dates' variables for 2 transportations between cities\\
\# Assert first transportation happens at first day (day 0), last transportation happens at last day (day 2)\\

\# Transportation methods \# \\
\# Set transportation method (flight, self-driving, taxi) variable for 2 transportations between cities\\
\# Assert only one of flight, self-driving, or taxi is used for 2 transportations between cities, self-driving is not valid if taxi or flight is used for any transportation\\
\# Assert all 2 transportations between cities are not flight\\

\# Flight information \# \\
\# Run FlightSearch to get flight info for Fort Lauderdale as origin, list of cities, city\_0, and dates ["2022-03-08", "2022-03-09", "2022-03-10"]\\
\# Get specific flight price info with Fort Lauderdale as origin and final destination, specific city, and departure date for 2 transportations\\
\# Set 'flight\_index' variable for 2 transportations\\
\# Assert 2 'flight\_index' variables are within valid range if taking flight, assert flight index to be -1 if not taking flight\\
\# Calculate flight price for 2 people for 2 transportations based on flight index variable\\
\# Get specific flight arrival time info with Fort Lauderdale as origin and final destination, specific city, and departure date for 2 transportations\\
\# Calculate flight arrival time for 2 transportations based on flight index variable\\

\# Driving information \# \\
\# Run DistanceSearch to get driving info for Fort Lauderdale as origin and city\_0\\
\# Get specific driving distance info with Fort Lauderdale as origin and final destination, specific city, and departure date for 2 transportations\\
\# Assert driving info is not empty if driving\\
\# Calculate self-driving and taxi price for 2 people and 2 transportations based on driving distance\\
\# Get driving arrival time with Fort Lauderdale as origin and final destination, specific city, and departure date for 2 transportations\\

\# Restaurant information \# \\
\# Get arrivals and city list for each day based on 2 transportations, 3 total travel day, and departure dates variables\\
\# Run RestaurantSearch to get restaurant price info and cuisine info for city\_0\\
\# Set 'restaurant\_in\_which\_city' variables for 9 (3 meals per day, 3 days) meals\\
\# For each 'restaurant\_in\_which\_city' variable, assert it to be either current city or next city based on transportation arrivals time\\
\# Set 'restaurant\_index' variables for 9 (3 meals per day, 3 days) meals\\
\# For each 'restaurant\_index', get specific price info based on 'restaurant\_in\_which\_city' variable, assert index are within valid range, assert restaurants in same city are not repeated, and calculate restaurant price for 2 people\\
\# Set 'cuisine\_type' variables for each cuisine type ["American", "Chinese"] required\\
\# For each cuisine type, iterate through all restaurant to check if it is satisfied\\

\# Attraction information \# \\
\# Run AttractionSearch to get attraction info for city\_0\\
\# Set 'attraction\_in\_which\_city' variables for 3 (1 per day) attractions\\
\# For each 'attraction\_in\_which\_city' variable, assert it to be either current city or next city based on transportation arrivals time\\
\# Set 'attraction\_index' variables for 3 (1 per day) attractions\\
\# For each 'attraction\_index', get specific length info based on attraction in which city variable, assert index are within valid range, and attrations in same city are not repeated\\

\# Accommodation information \# \\
\# Run AccommodationSearch to get accommodation info and accommodation constraints for city\_0\\
\# Set 'accommodation\_index' variables for 1 (1 per city) accommodations\\
\# For each 'accommodation\_index', get specific price info based on accommodation in which city variable, assert 'accommodation\_index' variable are within valid range, calculate number of room need for 2 people and accommodation price\\
\# For each city, get accommodation minimum night info and assert it to be less than the days stay in this city\\
\# For each 'accommodation\_index', get specific room type info, assert 'Private room' exist for all accommodations\\

\# Budget \# \\
\# Set budget limit variable to be 1100\\
\# Add 2 transportation price to spent, according to whether transportation method is flight, self-driving, or taxi\\
\# Add restaurant price to spent\\
\# Add accommodation price to spent\\
\# Assert current spent is within budget\\
-----EXAMPLES END-----\\
Based on the examples above, give the steps for following Natural Language query. Follow the original step structures.\\
Note to keep the format in examples and start each line containing steps with '\#' \\
Natural Language query:\\
\end{boxC}
\subsubsection{Step-Code prompt}
\label{sec:appendix-sat3}
The step to code example prompt for each constraint type is provided as follows: \\
Destination cities:\\
\vspace{-10pt}
\begin{minted}[fontsize=\scriptsize,breaklines]{python}
# Python script for testing satisfiability of the destination cities constraint of a travel plan problem. 

# Run CitySearch to get all possible destination cities in Minnesota State from origin 'Atlanta', remove origin 'Atlanta' if it is 
# in list
cities = CitySearch.run('Minnesota', 'Atlanta', query_json['date'])
if 'Atlanta' in cities:
        cities.remove('Atlanta')
# Set cities to be a list includes Milwaukee only
cities = ['Milwaukee']
# Loop through cities for 2 destination cities
for city_0_index, city_0 in enumerate(cities):
    for city_1_index, city_1 in enumerate(cities):
        # Initialize Z3 solver s
        s = Optimize()
        # Set 'city' variable to be indexes of 2 destination cities
        variables['city'] = [Int('city_' + str(i)) for i in range(2)]
        # If city_0_index and city_1_index are not same, assert 2 'city' variables equal to city index
        if city_0_index != city_1_index:
        s.assert_and_track(variables['city'][0] == city_0_index,  'visit city in cities list')
        s.assert_and_track(variables['city'][1] == city_1_index,  'visit city in cities list')
# Loop through cities for 1 destination cities
for city_0_index, city_0 in enumerate(cities):
    # Initialize Z3 solver s
    s = Optimize()
    # Set 'city' variable to be indexes of 1 destination cities
    variables['city'] = [Int('city_' + str(i)) for i in range(1)]
    # Assert 'city' variable equal to city index
    s.assert_and_track(variables['city'][0] == city_0_index,  'visit city in cities list')

# Based on the examples above, in which the lines start with '#' is the instuction, where the line/lines below it before the 
# next '#' is the corresponding code.
# For this below instruction, write corresponding code and respond instruction with code only. Start with ########## Destination 
# cities response########## and end with ########## Destination cities response ends##########.
\end{minted}
\vspace{10pt}
Departure Dates:\\
\vspace{-10pt}
\begin{minted}[fontsize=\scriptsize,breaklines]{python}
# Python script for testing satisfiability of the departure dates constraint of a travel plan problem. 

# Set 'departure_dates' variables for 3 transportations between cities
variables['departure_dates'] = [Int('departure_dates_transportation_' + str(i)) for i in range(3)]
# Assert first transportation happens at first day (day 0), last transportation happens at last day (day 6), second and third
# transportation happen in between but not at the same day
s.assert_and_track(variables['departure_dates'][0] == 0,  'travel start date')
s.assert_and_track(And(variables['departure_dates'][1] > 0, variables['departure_dates'][1] < variables['departure_dates'][2]),  'valid travel date')
s.assert_and_track(And(variables['departure_dates'][2] > variables['departure_dates'][1], variables['departure_dates'][1] < 6),  'valid travel date')
s.assert_and_track(variables['departure_dates'][3] == 6,  'travel end date')
# Assert first transportation happens at first day (day 0), last transportation happens at last day (day 2)
s.assert_and_track(variables['departure_dates'][0] == 0,  'travel start date')
s.assert_and_track(variables['departure_dates'][2] == 2,  'travel end date')

# Based on the examples above, in which the lines start with '#' is the instuction, where the line/lines below it before the next
# '#' is the corresponding code.
# For this below instruction, write corresponding code and respond instruction with code only. Start with ########## Departure 
# dates response########## and end with ########## Departure dates response ends##########.
\end{minted}
\vspace{10pt}
Transportation Methods:\\
\vspace{-10pt}
\begin{minted}[fontsize=\scriptsize,breaklines]{python}
# Python script for testing satisfiability of the transportation methods constraint of a travel plan problem. 

# Set transportation method variable (flight, self-driving, taxi) for 3 transportations between cities
variables['flight'] = [Bool('flight_travel_' + str(i)) for i in range(3)]
variables['self-driving'] = [Bool('self-driving_travel_' + str(i)) for i in range(3)]
variables['taxi'] = [Bool('taxi_travel_' + str(i)) for i in range(3)]
# Assert only one of flight, self-driving, or taxi is used for 3 transportations between cities, self-driving is not valid if taxi
# or flight is used for any transportation
s.assert_and_track(Or(variables['flight'][0], variables['self-driving'][0], variables['taxi'][0]),  'either flight, self-driving, or taxi for first transportation')
s.assert_and_track(Or(variables['flight'][1], variables['self-driving'][1], variables['taxi'][1]),  'either flight, self-driving, or taxi for second transportation')
s.assert_and_track(Or(variables['flight'][2], variables['self-driving'][2], variables['taxi'][2]),  'either flight, self-driving, or taxi for third transportation')
s.assert_and_track(Not(Or(And(variables['flight'][0], variables['self-driving'][0]), And(variables['flight'][0], variables['taxi'][0]), And(variables['taxi'][0], variables['self-driving'][0]))),  'flight, self-driving, and taxi not simutaneously for first transportation')
s.assert_and_track(Not(Or(And(variables['flight'][1], variables['self-driving'][1]), And(variables['flight'][1], variables['taxi'][1]), And(variables['taxi'][1], variables['self-driving'][1]))),  'flight, self-driving, and taxi not simutaneously for second transportation')
s.assert_and_track(Not(Or(And(variables['flight'][2], variables['self-driving'][2]), And(variables['flight'][2], variables['taxi'][2]), And(variables['taxi'][2], variables['self-driving'][2]))),  'flight, self-driving, and taxi not simutaneously for third transportation')
s.assert_and_track(Implies(Or(variables['flight'][0], variables['flight'][1], variables['flight'][2]), Not(Or(variables['self-driving'][0], variables['self-driving'][1], variables['self-driving'][2]))), 'no self-driving if taken flight for any transportation')
s.assert_and_track(Implies(Or(variables['taxi'][0], variables['taxi'][1], variables['taxi'][2]), Not(Or(variables['self-driving'][0], variables['self-driving'][1], variables['self-driving'][2]))), 'no self-driving if taken taxi for any transportation')
# Assert all 3 transportations between cities are not self-driving
s.assert_and_track(Not(variables['self-driving'][0]), 'no self-driving for first transportation')
s.assert_and_track(Not(variables['self-driving'][1]), 'no self-driving for second transportation')
s.assert_and_track(Not(variables['self-driving'][2]), 'no self-driving for third transportation')

# Based on the examples above, in which the lines start with '#' is the instuction, where the line/lines below it before the next
# '#' is the corresponding code.
# For this below instruction, write corresponding code and respond instruction with code only. Start with ########## Transportation
# response########## and end with ########## Transportation response ends##########.
\end{minted}
\vspace{10pt}
Flight Information:\\
\vspace{-10pt}
\begin{minted}[fontsize=\scriptsize,breaklines]{python}
# Python script for testing satisfiability of the flight constraint constraint of a travel plan problem. 

# Run FlightSearch to get flight info for Atlanta as origin, list of cities, city_0 and city_1, and dates
flight_info = FlightSearch.run_for_all_cities_and_dates('Atlanta', cities, [city_0, city_1], query_json['date'])
# Get specific flight price info with Atlanta as origin and final destination, specific city variable, and departure date for 3
# transportations
flight_0_price_list, flight_0_price_list_length = FlightSearch.get_info(flight_info, 'Atlanta', variables['city'][0], variables['departure_dates'][0], 'Price')
flight_1_price_list, flight_1_price_list_length = FlightSearch.get_info(flight_info, variables['city'][0], variables['city'][1], variables['departure_dates'][1], 'Price')
flight_2_price_list, flight_2_price_list_length = FlightSearch.get_info(flight_info, variables['city'][1], 'Atlanta', variables['departure_dates'][2], 'Price')
# Set 'flight_index' variable for 3 transportations
variables['flight_index'] = [Int('flight_{}_index'.format(i)) for i in range(3)]
# Assert 3 'flight_index' variables are within valid range if taking flight, assert flight index to be -1 if not taking flight
s.assert_and_track(Implies(variables['flight'][0], And(variables['flight_index'][0] >= 0,variables['flight_index'][0] < flight_0_price_list_length)), 'valid flight index for flight 0')
s.assert_and_track(Implies(variables['flight'][1], And(variables['flight_index'][1] >= 0,variables['flight_index'][1] < flight_1_price_list_length)), 'valid flight index for flight 1')
s.assert_and_track(Implies(variables['flight'][2], And(variables['flight_index'][2] >= 0,variables['flight_index'][2] < flight_2_price_list_length)), 'valid flight index for flight 2')
s.assert_and_track(Implies(Not(variables['flight'][0]), variables['flight_index'][0] == -1), 'valid flight index for flight 0')
s.assert_and_track(Implies(Not(variables['flight'][1]), variables['flight_index'][1] == -1), 'valid flight index for flight 1')
s.assert_and_track(Implies(Not(variables['flight'][2]), variables['flight_index'][2] == -1), 'valid flight index for flight 2')
# Calculate flight price for 2 people for 3 transportations based on flight index variable
flight_0_price = 2 * FlightSearch.get_info_for_index(flight_0_price_list, variables['flight_index'][0])
flight_1_price = 2 * FlightSearch.get_info_for_index(flight_1_price_list, variables['flight_index'][1])
flight_2_price = 2 * FlightSearch.get_info_for_index(flight_2_price_list, variables['flight_index'][2])
# Get specific flight arrival time info with Atlanta as origin and final destination, specific city, and departure date for 3
# transportations
flight_0_arrtime_list, _ = FlightSearch.get_info(flight_info, 'Atlanta', variables['city'][0], variables['departure_dates'][0], 'ArrTime')
flight_1_arrtime_list, _ = FlightSearch.get_info(flight_info, variables['city'][0], variables['city'][1], variables['departure_dates'][1], 'ArrTime')
flight_2_arrtime_list, _ = FlightSearch.get_info(flight_info, variables['city'][1], 'Atlanta', variables['departure_dates'][2], 'ArrTime')
# Calculate flight arrival time for 3 transportations based on flight index variable
flight_0_arrtime = FlightSearch.get_info_for_index(flight_0_arrtime_list, variables['flight_index'][0])
flight_1_arrtime = FlightSearch.get_info_for_index(flight_1_arrtime_list, variables['flight_index'][1])
flight_2_arrtime = FlightSearch.get_info_for_index(flight_2_arrtime_list, variables['flight_index'][2])

# Based on the examples above, in which the lines start with '#' is the instuction, where the line/lines below it before the next
# '#' is the corresponding code.
# For this below instruction, write corresponding code and respond instruction with code only. Start with ########## Flight
# response########## and end with ########## Flight response ends##########.

\end{minted}
\vspace{10pt}
Driving Information:\\
\vspace{-10pt}
\begin{minted}[fontsize=\scriptsize,breaklines]{python}
# Python script for testing satisfiability of the driving constraint of a travel plan problem. 

# Run DistanceSearch to get driving info for Atlanta as origin and city_0 and city_1
driving_info = DistanceSearch.run_for_all_cities('Atlanta', cities, [city_0, city_1])
# Get specific driving distance info with Atlanta as origin and final destination, specific city, and departure date for 3 
# transportations
driving_0_distance, driving_0_length = DistanceSearch.get_info(driving_info, 'Atlanta', variables['city'][0], 'Distance')
driving_1_distance, driving_1_length = DistanceSearch.get_info(driving_info, variables['city'][0], variables['city'][1], 'Distance')
driving_2_distance, driving_2_length = DistanceSearch.get_info(driving_info, variables['city'][1],'Atlanta', 'Distance')
# Assert driving info is not empty if driving
s.assert_and_track(Implies(Or(variables['self-driving'][0], variables['taxi'][0]), driving_0_length > 0), 'driving is possible for transportation 0')
s.assert_and_track(Implies(Or(variables['self-driving'][1], variables['taxi'][1]), driving_1_length > 0), 'driving is possible for transportation 1')
s.assert_and_track(Implies(Or(variables['self-driving'][2], variables['taxi'][2]), driving_2_length > 0), 'driving is possible for transportation 2')
# Calculate self-driving and taxi price for 3 people and 3 transportations based on driving distance
self_driving_0_price = 0.05 * driving_0_distance * math.ceil(3 / 5)
self_driving_1_price = 0.05 * driving_1_distance * math.ceil(3 / 5)
self_driving_2_price = 0.05 * driving_2_distance * math.ceil(3 / 5)
taxi_0_price = driving_0_distance * math.ceil(3 / 4)
taxi_1_price = driving_1_distance * math.ceil(3 / 4)
taxi_2_price = driving_2_distance * math.ceil(3 / 4)
# Get driving arrival time with Atlanta as origin and final destination, specific city, and departure date for 3 transportations
driving_0_arrtime, _ = DistanceSearch.get_info(driving_info, 'Atlanta', variables['city'][0], 'Duration')
driving_1_arrtime, _ = DistanceSearch.get_info(driving_info, variables['city'][0], variables['city'][1], 'Duration')
driving_2_arrtime, _ = DistanceSearch.get_info(driving_info, variables['city'][1], 'Atlanta', 'Duration')

# Based on the examples above, in which the lines start with '#' is the instuction, where the line/lines below it before the next
# '#' is the corresponding code.
# Follow the variable names in examples.
# For this below instruction, write corresponding code and respond instruction with code only. Start with ########## Driving
# response########## and end with ########## Driving response ends##########.
\end{minted}
\vspace{10pt}
Restaurant Information:\\
\vspace{-10pt}
\begin{minted}[fontsize=\scriptsize,breaklines]{python}
# Python script for testing satisfiability of the restaurant constraint of a travel plan problem. 

# Get arrivals and city list for each day based on 3 transportations, 5 total travel day, and departure dates variables
transportation_0_arrtime = If(variables['flight'][0], flight_0_arrtime, driving_0_arrtime)
transportation_1_arrtime = If(variables['flight'][1], flight_1_arrtime, driving_1_arrtime)
transportation_2_arrtime = If(variables['flight'][2], flight_2_arrtime, driving_2_arrtime)
arrives = get_arrivals_list([transportation_0_arrtime, transportation_1_arrtime, transportation_2_arrtime], 5, variables['departure_dates'])
city_list = get_city_list(variables['city'], 5, variables['departure_dates'])
# Run RestaurantSearch to get restaurant price info and cuisine info for city_0 and city_1
restaurant_price, restaurant_cuisines = RestaurantSearch.run_for_all_cities(cities, [city_0, city_1]) 
# Run RestaurantSearch to get restaurant price info and cuisine info for city_0
restaurant_price, restaurant_cuisines = RestaurantSearch.run_for_all_cities(cities, [city_0]) 
# Set 'restaurant_in_which_city' variables for 15 (3 meals per day, 5 days) meals
variables['restaurant_in_which_city'] = [Int('restaurant_' + str(i)) for i in range(3*5)]
# For each 'restaurant_in_which_city' variable, assert it to be either current city or next city based on transportation arrivals
# time
for i, variable in enumerate(variables['restaurant_in_which_city']):
    date_index = i // 3
    meal_index = i % 3
    if meal_index == 0: # breakfast
        s.assert_and_track(Or(variable == city_list[date_index], variable == city_list[date_index+1]),  'eat in which city b')
        s.assert_and_track(Implies(arrives[date_index]> 10, variable == city_list[date_index]),'eat in which city b')
        s.assert_and_track(Implies(arrives[date_index]< 5, variable == city_list[date_index+1]),'eat in which city b')
    if meal_index == 1: # lunch
        s.assert_and_track(Or(variable == city_list[date_index], variable == city_list[date_index+1]),  'eat in which city l')
        s.assert_and_track(Implies(arrives[date_index]> 15, variable == city_list[date_index]),'eat in which city l')
        s.assert_and_track(Implies(arrives[date_index]< 10, variable == city_list[date_index+1]),'eat in which city l')
    if meal_index == 2: # dinner
        s.assert_and_track(Or(variable == city_list[date_index], variable == city_list[date_index+1]),  'eat in which city d')
        s.assert_and_track(Implies(arrives[date_index]> 22, variable == city_list[date_index]),'eat in which city d')
        s.assert_and_track(Implies(arrives[date_index]< 17, variable == city_list[date_index+1]),'eat in which city d')
# Set 'restaurant_index' variables for 15 (3 meals per day, 5 days) meals
variables['restaurant_index'] = [Int('restaurant_{}_index'.format(i)) for i in range(3*5)]
# For each 'restaurant_index', get specific price info based on 'restaurant_in_which_city' variable, assert index are within valid
# range, assert restaurants in same city are not repeated, and calculate restaurant price for 2 people
all_restaurant_price = 0
for i, variable in enumerate(variables['restaurant_index']):
    restaurant_price_list, restaurant_list_length = RestaurantSearch.get_info(restaurant_price, variables['restaurant_in_which_city'][i], 'Price')
    s.assert_and_track(Implies(variables['restaurant_in_which_city'][i] != -1, And(variable >= 0, variable < restaurant_list_length)), 'valid restaurant index')
    s.assert_and_track(Implies(variables['restaurant_in_which_city'][i] == -1, variable == -1), 'valid restaurant index')
    for j in range(i-1, -1, -1):
        s.assert_and_track(Implies(And(variables['restaurant_in_which_city'][i] != -1, variables['restaurant_in_which_city'][i] == variables['restaurant_in_which_city'][j]), variable != variables['restaurant_index'][j]), 'non repeating restaurant index')
    Calculate restaurant price based on restaurant index
    all_restaurant_price += 2 * If(variables['restaurant_in_which_city'][i] != -1,  RestaurantSearch.get_info_for_index(restaurant_price_list, variable), 0)
# Set 'cuisine_type' variables for each cuisine type required
variables['cuisines_type'] = [Int('cuisines_' + i) for i in query_json['local_constraint']['cuisine']]
# For each cuisine type, iterate through all restaurant to check if it is satisfied
for index, cuisine in enumerate(query_json['local_constraint']['cuisine']):
    count = 0
    for i, variable in enumerate(variables['restaurant_index']):
        restaurant_cuisines_list, _ = RestaurantSearch.get_info(restaurant_cuisines, variables['restaurant_in_which_city'][i], 'Cuisines')
        count += If(RestaurantSearch.check_exists(cuisine, restaurant_cuisines_list, variable), 1, 0)
    s.assert_and_track(variables['cuisines_type'][index] == count,  cuisine + 'type restaurant')
    s.assert_and_track(variables['cuisines_type'][index] > 0,  cuisine + 'type restaurant is visited')

# Based on the examples above, in which the lines start with '#' is the instuction, where the line/lines below it before the next 
# '#' is the corresponding code.
# For this below instruction, write corresponding code and respond instruction with code only. Start with ########## Restaurant 
# response########## and end with ########## Restaurant response ends##########.
\end{minted}
\vspace{10pt}
Attraction Information:\\
\vspace{-10pt}
\begin{minted}[fontsize=\scriptsize,breaklines]{python}
# Python script for testing satisfiability of the attraction constraint of a travel plan problem. 

# Run AttractionSearch to get attraction info for city_0 and city_1
attraction_info = AttractionSearch.run_for_all_cities(cities, [city_0, city_1])
# Run AttractionSearch to get attraction info for city_0
attraction_info = AttractionSearch.run_for_all_cities(cities, [city_0])
# Set 'attraction_in_which_city' variables for 5 (1 per day) attractions
variables['attraction_in_which_city'] = [Int('attraction_' + str(i)) for i in range(1*5)]
# For each 'attraction_in_which_city' variable, assert it to be either current city or next city based on transportation arrivals
# time
for i, variable in enumerate(variables['attraction_in_which_city']):
    s.assert_and_track(variable == If(arrives[i]> 18, city_list[i], city_list[i+1]),  'attraction in which city')
# Set 'attraction_index' variables for 5 (1 per day) attractions
variables['attraction_index'] = [Int('attraction_{}_index'.format(i)) for i in range(1*5)]
# For each 'attraction_index', get specific length info based on attraction in which city variable, assert index are within valid
# range, and attrations in same city are not repeated
for i, variable in enumerate(variables['attraction_index']):
    attraction_list_length = AttractionSearch.get_info(attraction_info, variables['attraction_in_which_city'][i])
    s.assert_and_track(Implies(variables['attraction_in_which_city'][i] != -1, And(variable >= 0, variable < attraction_list_length)), 'valid attraction index')
    s.assert_and_track(Implies(variables['attraction_in_which_city'][i] == -1, variable == -1), 'valid attraction index')
    for j in range(i-1, -1, -1):
        s.assert_and_track(Implies(And(variables['attraction_in_which_city'][i] != -1, variables['attraction_in_which_city'][i] == variables['attraction_in_which_city'][j]), variable != variables['attraction_index'][j]), 'non repeating attraction index')

# Based on the examples above, in which the lines start with '#' is the instuction, where the line/lines below it before the next 
# '#' is the corresponding code.
# For this below instruction, write corresponding code and respond instruction with code only. Start with ########## Attraction 
# response########## and end with ########## Attraction response ends##########.

\end{minted}
\vspace{10pt}
Accommodation Information:\\
\vspace{-10pt}
\begin{minted}[fontsize=\scriptsize,breaklines]{python}
# Python script for testing satisfiability of the accommodation constraint of a travel plan problem. 

# Run AccommodationSearch to get accommodation info and accommodation constraints for city_0 and city_1
accommodation_info, accommodation_constraints = AccommodationSearch.run_for_all_cities(cities, [city_0, city_1]) 
# Run AccommodationSearch to get accommodation info and accommodation constraints for city_0
accommodation_info, accommodation_constraints = AccommodationSearch.run_for_all_cities(cities, [city_0]) 
# Set 'accommodation_index' variables for 2 (1 per city) accommodations
variables['accommodation_index'] = [Int('accommodation_{}_index'.format(i)) for i in range(2)]
# For each 'accommodation_index', get specific price info based on accommodation in which city variable, assert 
# 'accommodation_index' variable are within valid range, calculate number of room need for 2 people and accommodation price
all_accommodation_price = 0
for i, variable in enumerate(variables['accommodation_index']):
    accommodation_price_list, accommodation_list_length = AccommodationSearch.get_info(accommodation_info, variables['city'][i], 'Price')
    s.assert_and_track(And(variable >= 0, variable < accommodation_list_length), 'valid accomodation index')
    accommodation_maximum_occupancy_list, _ = AccommodationSearch.get_info(accommodation_info, variables['city'][i], 'Maximum_occupancy')
    num_room = convert_to_int(RealVal(2) / AccommodationSearch.get_info_for_index(accommodation_maximum_occupancy_list, variable))
    all_accommodation_price += (variables['departure_dates'][i+1] - variables['departure_dates'][i]) * num_room * AccommodationSearch.get_info_for_index(accommodation_price_list, variable)
# For each city, get accommodation minimum night info and assert it to be less than the days stay in this city
for index, city in enumerate(variables['city']):
    accommodation_minimum_nights_list, _ = AccommodationSearch.get_info(accommodation_info, city, 'Minimum_nights')
    minimum_night = AccommodationSearch.get_info_for_index(accommodation_minimum_nights_list, variables['accommodation_index'][index])
    s.assert_and_track(minimum_night <= variables['departure_dates'][index+1]- variables['departure_dates'][index], 'minimum nights satisfied')
# For each 'accommodation_index', get specific room type and house rules info, assert 'Entire home/apt' exist for all
# accommodations, assert 'No parties' does not exist for all accommodations
for i, variable in enumerate(variables['accommodation_index']):
        accommodation_room_types_list, _ = AccommodationSearch.get_info(accommodation_constraints, variables['city'][i], 'Room_types')
        accommodation_house_rules_list, _ = AccommodationSearch.get_info(accommodation_constraints, variables['city'][i], 'House_rules')
        s.assert_and_track(AccommodationSearch.check_exists('Entire home/apt', accommodation_room_types_list, variable) == True,  'Entire home/apt' + 'types accomadation visited')
        s.assert_and_track(AccommodationSearch.check_exists('No parties', accommodation_house_rules_list, variable) == False,  'No parties' + 'rules accomadation not visited')

# Based on the examples above, in which the lines start with '#' is the instuction, where the line/lines below it before the next
# '#' is the corresponding code.
# For this below instruction, write corresponding code and respond instruction with code only. Start with ########## Accommodation
# response########## and end with ########## Accommodation response ends##########.
\end{minted}
\vspace{10pt}
Budget:\\
\vspace{-10pt}
\begin{minted}[fontsize=\scriptsize,breaklines]{python}
# Python script for testing satisfiability of the budget constraint of a travel plan problem. 

# Set budget limit variable to be 7900
variables['budget_limit'] = RealVal(7900)
# Add 3 transportation price to spent, according to whether transportation method is flight, self-driving, or taxi
spent = 0
spent += If(variables['flight'][0], flight_0_price, If(variables['self-driving'][0], self_driving_0_price, If(variables['taxi'][0], taxi_0_price, 10000)))
spent += If(variables['flight'][1], flight_1_price, If(variables['self-driving'][1], self_driving_1_price, If(variables['taxi'][1], taxi_1_price, 10000)))
spent += If(variables['flight'][2], flight_2_price, If(variables['self-driving'][2], self_driving_2_price, If(variables['taxi'][2], taxi_2_price, 10000)))
# Add restaurant price to spent
spent += all_restaurant_price
# Add accommodation price to spent
spent += all_accommodation_price
# Assert current spent is within budget
s.assert_and_track(spent <= variables['budget_limit'], 'budget enough')

# Based on the examples above, in which the lines start with '#' is the instuction, where the line/lines below it before the next 
# '#' is the corresponding code.
# For this below instruction, write corresponding code and respond instruction with code only. Start with ########## Budget 
# response########## and end with ########## Budget response ends##########.
\end{minted}
\subsubsection{Prompt difference of GPT-4, Claude 3, and Mistral-Large}
\label{sec:appendix-sat4}
With the prompt we have for GPT-4 as the starting point, we adjust the prompts (add more explainations or examples) for Claude-3 and Mistral-Large using the training set in TravelPlanner. \\
\\
Claude-3 almost has the same prompt as GPT-4, except for the JSON-Step prompt. Since in training set, a failure case for Claude-3 is it is not able to handle the "house rule" properly. When the JSON specifies "house rule" to be "children under 10" it means the travellers have children under 10 and would like to stay in accommodations without "No children under 10" rule. While Claude-3 sometimes is not able to give "No children under 10" in the step, instead, it gives steps with "children under 10 not allowed". To enable it to handle this, we add one sentence explanation \textbf{"if house rule 'xxx' is mentioned, then 'No xxx' should not exist for all accomadations."} in JSON-Step prompt.\\
\\
Compared to Claude-3, Mistral-Large needs more prompt adjustment: 
\begin{itemize}
\setlength\itemsep{-0.3em}
\item We add \textbf{"You can only assign null to local constraints if it is needed. Other fields must have values."} to NL-JSON prompt because Mistral-Large sometimes misses some information in JSON translation.
\item Claude-3 uses the same JSON-Step prompt as Mistral-Large. 
\item We add a 3-city loop-through-cities example in Destination Cities Step-Code prompt; We add a 2-city travel-date-assertion example in Departure Dates prompt; We add a 1-city transportation-method assertion-example to Transportation Methods; We add instructions that ask LLM to not use for-loops and name variable with "i" as when it tries to iteratively create or access variables with i it fails to write the correct code. 
\end{itemize}
From the amount of changes we need to make, we can observe that Mistral-Large in general produces more code generation errors compared to GPT-4 and Claude-3, thus needing more examples and explanations provided in prompts.

\subsubsection{Failure cases of Mistral-Large}
\label{sec:appendix-sat5}
Although we tune our prompt with training set, there are still some failure cases that do not appear in training set and thus negatively affect Mistral-Large's delivery rate.\\
The major failure mode is: "room type" takes the value "no shared room". This room type is special in that when other room types such as "private room" is specified, the generated instruction steps should be "private room exists for all accommodations". However, only when "no shared room" is mentioned, the steps should be "shared room does not exist for all accommodations." Since "no shared room" does not appear in training set or examples, and Mistral-Large is not able to generalize to it, it fails by producing "not shard room exists for all accommodations", thus fails to search for home with type "not shared room". This is the major failure mode and is responsible for 7.8\% of failed delivered plans (15.0\% in total). Other than this, the failures are induced by runtime issue or some occasional code generation errors. 

\subsection{Prompts for Interactive Plan Repair}
\label{sec:appendix-unsat}
\subsubsection{Suggestion prompt}
The instruction prompt that guides LLM to collect information, analyze current situation, and offer suggestions for unsatisfiable queries is provided as follows: \\
Suggestion prompt for UnsatChristmas:\\
\vspace{-15pt}
\begin{boxC}
As a travel planner, you have some constraints in JSON format to satisfy for a round trip travel plan problem. \\
The trip spans "date", goes from "org", travels "dest" cities in a row in between, and then goes back to "org"\\
For example, if "org" is city\_0, and "dest" is [city\_1, city\_2], then the flights could be [city\_0->city\_1,city\_1->city\_2,city\_2->city\_0] or [city\_0->city\_2,city\_2-> city\_1,city\_1->city\_0] \\
In addition, "local\_constraint" contains three possible constraints. "flight rule" specifies whether "non\-stop" is required or not. "airlines" specifies a list of a airlines user can accept. Possible options of "attraction\_category" specifies a list of categories of attractions want to visit. If the field value is null in JSON, this specific hard constraint is not included. \\
The specified "flight rule" needs to be satisfied by all flights. All flights need to be one of the accepted "airlines". All categories in "attraction\_category" needs to be satisfied, and a category could be satisfied if it exists for one attraction. \\
After anaylzed these constraints, you found they are not satisfiable under current setting. You will be giving unsatisfiable reasons.\\
Collect information based on the reasons or, based on the information you collect, analyze current situation or give a suggested modification to the constraints. \\
Info collecting can take 4 different actions: \\
(1) FlightSearch[Departure City, Destination City]:
Description: A flight information retrieval tool.
Parameters:\\
    Departure City: The city you'll be flying out from.\\
    Destination City: The city you aim to reach.\\
Example: FlightSearch[New York City, London] would fetch flights from New York City to London.\\
\\
(2) AirlineSearch[Airline]:\\
Description: Find flights of input airline.\\
Parameter: Airline - The airline name you want to take. \\
Example: AirlineSearch[United] would return all flights of United airline.\\
\\
(3) AttractionSearch[City]:\\
Description: Find attractions categories in a city of your choice.\\
Parameter: City - The name of the city where you're seeking attractions. \\
Example: AttractionSearch[London] would return attraction categories in London.\\
\\
(4) CategorySearch[Category]:\\
Description: Find cities contain attractions of input category.\\
Parameter: Category - The attraction category where you want to visit. \\
Example: CategorySearch[Park] would return all cities where attractions of category Park exist.\\
\\
You need to take an action analyze current situation and plan your future steps after each FlightSearch, AirlineSearch, AttractionSearch, or CategorySearch.\\
Example: Analyze[your analysis of current situation and plan for future]\\

You can suggest to remove the non-stop constraint, suggest to change required airlines, suggest to change destination cities(but keep number of destination cities unchanged), suggest to change attraction categories, or suggest to raise budget. 
Do not give other suggestions that change other fields in JSON input, such as origin, number of visit cities, etc.\\
Please give a reasonable suggestion to modify the constraint only when you think you've collected enough information and the suggestion has high chance to be satisfiable. For example, if destination city does not have required attraction category, you should suggest to change destination city if info shows the new city has the required category\\
Please try to keep original constraint and make minimal change to original constraint only when it is necessary.\\
Examples: \\
    Suggest[raise budget to 5000]\\
    Suggest[change destination cities to be Istanbul and Macau]\\
    Suggest[remove the non-stop constraint]\\
    Suggest[change airlines to be United, Air France, or JetBlue]\\
    Suggest[change attraction categories to be Garden and Museum]\\
\\
A list of possible cities is ['Bangkok', 'Dubai', 'Hong Kong', 'Istanbul', 'Kuala Lumpur', 'London', 'Macau', 'New York City', 'Paris', 'Singapore']\\
Now, based on the input query, unsatisfiable reasons, and collected information, please give the next action(only one action) you want to take only with no explainations, you need to give a suggestion within 15 iterations: \\
Input query: \{query\} \\
Unsatisfiable reasons: \{reasons\}\\
Collected information: \{info\}
\end{boxC}

Suggestion prompt for TravelPlanner:\\
\vspace{-15pt}
\begin{boxC}
As a travel planner, you have some constraints in JSON format to satisfy for a round trip travel plan problem. \\
The trip spans "date", goes from "org", travels "dest" city, and then goes back to "org"
For example, if "org" is city\_0, and "dest" is city\_1, then the transportations would be [city\_0->city\_1,city\_1->city\_0]\\
In addition, "local\_constraint" contains four possible constraints. Possible options of "house rule" includes ["parties", "smoking", "children under 10", "pets", "visitors"]. Possible options of "cuisine" includes ["Chinese", "American", "Italian", "Mexican", "Indian", "Mediterranean", "French"]. Possible options of "house type" includes ["entire room", "private room", "shared room", "not shared room"]. Possible options of "transportation" includes ["no flight", "no self-driving"]. If the field value is null in JSON, this specific hard constraint is not included.\\ 
The specified "house rule" and "house type" needs to be satisfied by all accommodations. The specified "transportation" needs to be satisfied by all transportations. All cuisines in "cuisine" needs to be satisfied, and a cuisine could be satisfied if it exists for one restaurant. \\
After anaylzed these constraints, you found they are not satisfiable under current setting. You will be giving unsatisfiable reasons.\\
Collect information based on the reasons or, based on the information you collect, analyze current situation or give a suggested modification to the constraints. \\
Info collecting can take 6 different actions: \\
(1) DrivingCheck[Departure City, Destination City]:
Description: A driving information checking tool that checks if driving is feasible.\\
Parameters:\\
    Departure City: The city you'll be driving out from.\\
    Destination City: The city you aim to reach.\\
Example: DrivingCheck[Grand Forks, Minneapolis] would check if driving is feasible from Grand Forks to Minneapolis.\\
\\
(2) DrivingSearch[Departure City]:\\
Description: A driving information retrieval tool that returns all reachable cities.\\
Parameters:\\
    Departure City: The city you'll be driving out from.\\
Example: DrivingSearch[Grand Forks] would return all reachable cities from Grand Forks through driving.\\
\\
(3) FlightCheck[Departure City, Destination City, Date]:\\
Description: A flight information checking tool that checks if flight is feasible.\\
Parameters:\\
    Departure City: The city you'll be flying out from.\\
    Destination City: The city you aim to reach.\\
    Date: The date you take the flight.\\
Example: FlightCheck[Grand Forks, Minneapolis, 2022-06-05] would check if flight is feasible from Grand Forks to Minneapolis on 2022-06-05.\\
\\
(4) FlightSearch[Departure City, Date]:
Description: A flight information retrieval tool that returns all reachable cities.\\
Parameters:\\
    Departure City: The city you'll be flying out from.\\
    Date: The date you take the flight.\\
Example: FlightSearch[Grand Forks, 2022-06-05] would return all reachable cities from Grand Forks through flight on 2022-06-05.\\
\\
(5) AccommodationSearch[City]:\\
Description: Find accommodations types in a city of your choice.\\
Parameter: City - The name of the city where you're seeking accommodations. \\
Example: AccommodationSearch[Grand Forks] would return accommodation categories in Grand Forks.\\
\\
(6) TypeSearch[Type]:\\
Description: Find cities contain accommodations of input type.\\
Parameter: Type - The accommodation type where you want to visit. \\
Example: TypeSearch[entire room] would return all cities where entire room type accommodation exist.\\
\\
You need to take an action analyze current situation and plan your future steps after each DrivingCheck, DrivingSearch, FlightCheck, FlightSearch, AccommodationSearch, or TypeSearch.\\
Example: Analyze[your analysis of current situation and plan for future]\\

You can suggest to remove the "house type" constraint, suggest to remove the "transportation" constraint, suggest to change destination cities(but keep number of destination cities unchanged), or suggest to raise budget. 
Do not give other suggestions that change other fields in JSON input, such as origin, number of visit cities, etc.\\
Please give a reasonable suggestion to modify the constraint only when you think you've collected enough information and the suggestion has high chance to be satisfiable. For example, if destination city does not have required accomadation type, you should suggest to change destination city if info shows the new city has the required type.\\
Please try to keep original constraint and make minimal change to original constraint only when it is necessary.\\
Examples: \\
    Suggest[raise budget to 5000]\\
    Suggest[change destination cities to be Minneapolis]\\
    Suggest[remove the house type constraint]\\
    Suggest[remove the flight/no flight/ no self-driving assertion for transportations]\\
\\
Now, based on the input query, unsatisfiable reasons, and collected information, please give the next action(only one action) you want to take only with no explainations, you need to give a suggestion within 15 iterations: \\
Input query: \{query\} \\
Unsatisfiable reasons: \{reasons\}\\
Collected information: \{info\}
\end{boxC}
\subsubsection{Suggestion-No Reason prompt}
The Suggestion-No Reason prompt is basically modified from the Suggestion prompt by removing all descriptions about reasons.
\subsubsection{Suggestion-No Sovler prompt}
The instruction suggestion prompt that remove the iterative solver calling and directly guide LLM to offer a list of suggestions is provided as follows: \\
Suggestion-No Sovler prompt for UnsatChristmas:\\
\vspace{-15pt}
\begin{boxC}
As a travel planner, you have some constraints in JSON format to satisfy for a round trip travel plan problem. \\
The trip spans "date", goes from "org", travels "dest" cities in a row in between, and then goes back to "org"
For example, if "org" is city\_0, and "dest" is [city\_1, city\_2], then the flights could be [city\_0->city\_1,city\_1->city\_2,city\_2->city\_0] or [city\_0->city\_2,city\_2-> city\_1,city\_1->city\_0] \\
In addition, "local\_constraint" contains three possible constraints. "flight rule" specifies whether "non-stop" is required or not. "airlines" specifies a list of a airlines user can accept. Possible options of "attraction\_category" specifies a list of categories of attractions want to visit. If the field value is null in JSON, this specific hard constraint is not included. \\
The specified "flight rule" needs to be satisfied by all flights. All flights need to be one of the accepted "airlines". All categories in "attraction\_category" needs to be satisfied, and a category could be satisfied if it exists for one attraction. \\
After anaylzed these constraints, you found they are not satisfiable under current setting.\\
Collect information or, based on the information you collect, analyze current situation or give a suggested modification to the constraints. \\
Info collecting can take 4 different actions: \\
(1) FlightSearch[Departure City, Destination City]:\\
Description: A flight information retrieval tool.
Parameters:\\
    Departure City: The city you'll be flying out from.\\
    Destination City: The city you aim to reach.\\
Example: FlightSearch[New York City, London] would fetch flights from New York City to London.\\
\\
(2) AirlineSearch[Airline]:\\
Description: Find flights of input airline.\\
Parameter: Airline - The airline name you want to take. \\
Example: AirlineSearch[United] would return all flights of United airline.\\
\\
(3) AttractionSearch[City]:\\
Description: Find attractions categories in a city of your choice.\\
Parameter: City - The name of the city where you're seeking attractions. \\
Example: AttractionSearch[London] would return attraction categories in London.\\
\\
(4) CategorySearch[Category]:\\
Description: Find cities contain attractions of input category.\\
Parameter: Category - The attraction category where you want to visit. \\
Example: CategorySearch[Park] would return all cities where attractions of category Park exist.\\
\\
You need to take an action analyze current situation and plan your future steps after each FlightSearch, AirlineSearch, AttractionSearch, or CategorySearch.\\
Example: Analyze[your analysis of current situation and plan for future]\\
\\
You can suggest to remove the non-stop constraint, suggest to change required airlines, suggest to change destination cities(but keep number of destination cities unchanged), suggest to change attraction categories, or suggest to raise budget. 
Do not give other suggestions that change other fields in JSON input, such as origin, number of visit cities, etc.\\
Please give reasonable suggestions to modify the constraint only when you think you've collected enough information and the suggestion has high chance to be satisfiable. For example, if destination city does not have required attraction category, you should suggest to change destination city if info shows the new city has the required category\\
Please try to keep original constraint and make minimal change to original constraint only when it is necessary.\\
You can give one or more suggestions if you think one is not enough. Please separate the suggestions with ;\\
Examples:  \\
    Suggest[raise budget to 5000]\\
    Suggest[change destination cities to be Istanbul and Macau]\\
    Suggest[remove the non-stop constraint]\\
    Suggest[change airlines to be United, Air France, or JetBlue]\\
    Suggest[change attraction categories to be Garden and Museum]\\
    Suggest[raise budget to 3000; change destination cities to be London]\\
\\
A list of possible cities is ['Bangkok', 'Dubai', 'Hong Kong', 'Istanbul', 'Kuala Lumpur', 'London', 'Macau', 'New York City', 'Paris', 'Singapore']
Now, based on the input query and collected information, please give the next action(only one action) you want to take only with no explainations, you need to give suggestions within 15 iterations: \\
Input query: \{query\} \\
Collected information: \{info\}
\end{boxC}
Suggestion-No Sovler prompt for UnsatChristmas:\\
\vspace{-15pt}
\begin{boxC}
As a travel planner, you have some constraints in JSON format to satisfy for a round trip travel plan problem. \\
The trip spans "date", goes from "org", travels "dest" city, and then goes back to "org"
For example, if "org" is city\_0, and "dest" is city\_1, then the transportations would be [city\_0->city\_1,city\_1->city\_0]\\
In addition, "local\_constraint" contains four possible constraints. Possible options of "house rule" includes ["parties", "smoking", "children under 10", "pets", "visitors"]. Possible options of "cuisine" includes ["Chinese", "American", "Italian", "Mexican", "Indian", "Mediterranean", "French"]. Possible options of "house type" includes ["entire room", "private room", "shared room", "not shared room"]. Possible options of "transportation" includes ["no flight", "no self-driving"]. If the field value is null in JSON, this specific hard constraint is not included. 
The specified "house rule" and "house type" needs to be satisfied by all accommodations. The specified "transportation" needs to be satisfied by all transportations. All cuisines in "cuisine" needs to be satisfied, and a cuisine could be satisfied if it exists for one restaurant. \\
After anaylzed these constraints, you found they are not satisfiable under current setting.\\
Collect information or, based on the information you collect, analyze current situation or give a suggested modification to the constraints.\\ 
Info collecting can take 6 different actions: \\
(1) DrivingCheck[Departure City, Destination City]:
Description: A driving information checking tool that checks if driving is feasible.\\
Parameters:\\
    Departure City: The city you'll be driving out from.\\
    Destination City: The city you aim to reach.\\
Example: DrivingCheck[Grand Forks, Minneapolis]\\ would check if driving is feasible from Grand Forks to Minneapolis.\\
\\
(2) DrivingSearch[Departure City]:\\
Description: A driving information retrieval tool that returns all reachable cities.\\
Parameters:\\
    Departure City: The city you'll be driving out from.\\
Example: DrivingSearch[Grand Forks] would return all reachable cities from Grand Forks through driving.\\
\\
(3) FlightCheck[Departure City, Destination City, Date]:\\
Description: A flight information checking tool that checks if flight is feasible.\\
Parameters:\\
    Departure City: The city you'll be flying out from.\\
    Destination City: The city you aim to reach.
    Date: The date you take the flight.\\
Example: FlightCheck[Grand Forks, Minneapolis, 2022-06-05] would check if flight is feasible from Grand Forks to Minneapolis on 2022-06-05.\\
\\
(4) FlightSearch[Departure City, Date]:\\
Description: A flight information retrieval tool that returns all reachable cities.\\
Parameters:\\
    Departure City: The city you'll be flying out from.\\
    Date: The date you take the flight.\\
Example: FlightSearch[Grand Forks, 2022-06-05] would return all reachable cities from Grand Forks through flight on 2022-06-05.\\
\\
(5) AccommodationSearch[City]:\\
Description: Find accommodations types in a city of your choice.\\
Parameter: City - The name of the city where you're seeking accommodations. \\
Example: AccommodationSearch[Grand Forks] would return accommodation categories in Grand Forks.\\
\\
(6) TypeSearch[Type]:\\
Description: Find cities contain accommodations of input type.\\
Parameter: Type - The accommodation type where you want to visit. \\
Example: TypeSearch[entire room] would return all cities where entire room type accommodation exist.
\\
You need to take an action analyze current situation and plan your future steps after each DrivingCheck, DrivingSearch, FlightCheck, FlightSearch, AccommodationSearch, or TypeSearch.\\
Example: Analyze[your analysis of current situation and plan for future]\\
\\
You can suggest to remove the "house type" constraint, suggest to remove the "transportation" constraint, suggest to change destination cities(but keep number of destination cities unchanged), or suggest to raise budget. 
Do not give other suggestions that change other fields in JSON input, such as origin, number of visit cities, etc.\\
Please give a reasonable suggestion to modify the constraint only when you think you've collected enough information and the suggestion has high chance to be satisfiable. For example, if destination city does not have required accomadation type, you should suggest to change destination city if info shows the new city has the required type.\\
Please try to keep original constraint and make minimal change to original constraint only when it is necessary.\\
You can give one or more suggestions if you think one is not enough. Please separate the suggestions with ;\\
Examples: \\
    Suggest[raise budget to 5000]\\
    Suggest[change destination cities to be Minneapolis]\\
    Suggest[remove the house type constraint]\\
    Suggest[remove the flight/no flight/ no self-driving assertion for transportations]\\
    Suggest[raise budget to 2000, change destination cities to be Chicago]\\
\\
Now, based on the input query and collected information, please give the next action(only one action) you want to take only with no explainations, you need to give a suggestion within 15 iterations: \\
Input query: \{query\} \\
Collected information: \{info\}
\end{boxC}
\subsubsection{Code modify prompt}
\begin{boxC}
As a travel planner, you have some python codes that tests the satisfiability of a travel plan problem.\\
While now some of the constraints are changed, your task is to change the python codes according to the changed constraints. \\
Only change the part of code that needs to be modified, and do not add any new parts.\\
Please respond with codes only, and be sure to include full codes instead of lines of updated codes.\\
Start with \#\#\#\#\#\#\#\#\#\# response \#\#\#\#\#\#\#\#\#\# and end with \#\#\#\#\#\#\#\#\#\# response ends \#\#\#\#\#\#\#\#\#\#.\\
\\
Original Codes: \{codes\}\\
Modified Constraints: \{constraints\}
\end{boxC}

\subsection{Prompts for Generalization Evaluation}
\label{sec:appendix-gen}
To test the capability of our framework to generalize to unseen constraint types, we add this task description and append at the end of Step-Code prompt: 
\begin{boxC}
Based on the examples above, can you give the steps for following JSON constraint with different fields.
In the input JSON, "org" denotes the departure city. "dest" denotes the destination city/cities. "days" denotes the total number of travel days. When "days" equals 5 or 7. "date" includes the specific date to visit.\\
In addition, "local\_constraint" contains three possible constraints. "flight rule" specifies the whether "non-stop" is required or not. "airlines" specifies a list of a airlines you can accept. Possible options of "attraction\_category" specifies a list of categories of attractions want to visit. "transportation" is always 'flight'. If the field value is null in JSON, this specific hard constraint is not included. \\
The specified "flight rule" needs to be satisfied by all flights. All flights need to be one of the accepted "airlines". All categories in "attraction\_category" needs to be satisfied, and a category could be satisfied if it exists for one attraction.\\
We do not consider restaurant information or accommodation information, instead, we fix each restaurant price to be 30 per person and accommodation to be 100 per person.
\end{boxC}
\subsection{Paraphrased Prompt Examples}
\label{sec:appendix-paraphrase}
Here's an example paraphrased prompt for Query-Step:
\begin{boxC}
You are provided with a natural language query that outlines specific constraints for a travel itinerary problem. These constraints may include details such as the departure city, destination state or city, total travel days and dates, and particular requirements related to accommodation, dining, and transportation. 

Your task is to provide a comprehensive, step-by-step guide to translate these constraints into code. Below are some example steps for various constraints:

-----EXAMPLE 1-----\\
Natural Language query:\\
Can you create a 5-day travel itinerary for a group of 3, departing from Atlanta and visiting 2 cities in Minnesota from March 3rd to March 7th, 2022? We have a budget of \$7,900. We require accommodations that allow parties and should ideally be entire rooms. Although we don't plan to self-drive, we would like the flexibility to host parties.\\
Steps:\\
\# Destination cities \# \\
\# Execute CitySearch to identify all potential destination cities in Minnesota State for the dates ["2022-03-16", "2022-03-17", "2022-03-18"] from the starting point 'Atlanta', and exclude 'Atlanta' if it appears in the list\\
\# Iterate over cities to select 2 destination cities\\
\# Initialize Z3 solver s\\
\# Define 'city' variable to represent indexes of 2 destination cities\\
\# Ensure city\_0\_index and city\_1\_index are distinct, and assert 2 'city' variables equal to city index\\

\# Departure dates \# \\
\# Define 'departure\_dates' variables for 3 transportations between cities\\
\# Assert that the first transportation occurs on the first day (day 0), the last transportation on the last day (day 4), and the second transportation can occur on any day in between\\

\# Transportation methods \# \\
\# Define transportation method (flight, self-driving, taxi) variable for 3 transportations between cities\\
\# Assert that only one of flight, self-driving, or taxi is used for 3 transportations between cities, and self-driving is invalid if taxi or flight is used for any transportation\\
\# Assert that all 3 transportations between cities are not self-driving\\

\# Flight information \# \\
\# Execute FlightSearch to obtain flight info for Atlanta as the origin, list of cities, city\_0 and city\_1, and dates ["2022-03-16", "2022-03-17", "2022-03-18"]\\
\# Retrieve specific flight price info with Atlanta as the origin and final destination, specific city variable, and departure date for 3 transportations\\
\# Define 'flight\_index' variable for 3 transportations\\
\# Assert that 3 'flight\_index' variables are within a valid range if taking a flight, and set flight index to -1 if not taking a flight\\
\# Calculate flight price for 3 people for 3 transportations based on flight index variable\\
\# Retrieve specific flight arrival time info with Atlanta as the origin and final destination, specific city, and departure date for 3 transportations\\
\# Calculate flight arrival time for 3 transportations based on flight index variable\\

\# Driving information \# \\
\# Execute DistanceSearch to obtain driving info for Atlanta as the origin and city\_0 and city\_1\\
\# Retrieve specific driving distance info with Atlanta as the origin and final destination, specific city, and departure date for 3 transportations\\
\# Assert that driving info is not empty if driving\\
\# Calculate self-driving and taxi price for 3 people and 3 transportations based on driving distance\\
\# Retrieve driving arrival time with Atlanta as the origin and final destination, specific city, and departure date for 3 transportations\\

\# Restaurant information \# \\
\# Obtain arrivals and city list for each day based on 3 transportations, 5 total travel days, and departure dates variables\\
\# Execute RestaurantSearch to obtain restaurant price info and cuisine info for city\_0 and city\_1\\
\# Define 'restaurant\_in\_which\_city' variables for 15 (3 meals per day, 5 days) meals\\
\# For each 'restaurant\_in\_which\_city' variable, assert it to be either the current city or the next city based on transportation arrivals time\\
\# Define 'restaurant\_index' variables for 15 (3 meals per day, 5 days) meals\\
\# For each 'restaurant\_index', retrieve specific price info based on 'restaurant\_in\_which\_city' variable, assert indexes are within a valid range, assert restaurants in the same city are not repeated, and calculate restaurant price for 3 people\\
\# Calculate restaurant price based on restaurant index\\

\# Attraction information \# \\
\# Execute AttractionSearch to obtain attraction info for city\_0 and city\_1\\
\# Define 'attraction\_in\_which\_city' variables for 5 (1 per day) attractions\\
\# For each 'attraction\_in\_which\_city' variable, assert it to be either the current city or the next city based on transportation arrivals time\\
\# Define 'attraction\_index' variables for 5 (1 per day) attractions\\
\# For each 'attraction\_index', retrieve specific length info based on attraction in which city variable, assert indexes are within a valid range, and attractions in the same city are not repeated\\

\# Accommodation information \# \\
\# Execute AccommodationSearch to obtain accommodation info and accommodation constraints for city\_0 and city\_1\\
\# Define 'accommodation\_index' variables for 2 (1 per city) accommodations\\
\# For each 'accommodation\_index', retrieve specific price info based on accommodation in which city variable, assert 'accommodation\_index' variable is within a valid range, calculate the number of rooms needed for 3 people and accommodation price\\
\# For each city, obtain accommodation minimum night info and assert it to be less than the days stayed in this city\\
\# For each 'accommodation\_index', retrieve specific room type and house rules info, assert 'Entire home/apt' exists for all accommodations, assert 'No parties' does not exist for all accommodations\\

\# Budget \# \\
\# Set budget limit variable to be 7900\\
\# Add 3 transportation prices to spent, according to whether the transportation method is flight, self-driving, or taxi\\
\# Add restaurant price to spent\\
\# Add accommodation price to spent\\
\# Assert current spent is within budget\\

-----EXAMPLE 2-----\\
Natural Language query:\\
Can you help with generating a 7-day travel plan for a party of 5? We're setting off from Indianapolis and planning to explore 3 cities in Colorado from March 11th to March 17th, 2022. We have a budget of \$15,100 for this trip. We'll be bringing our pets, so pet-friendly accommodations are a must. It's important for us to stay in accommodations that permit children under the age of 10. We're also hoping to find places that offer Mexican, Italian, Mediterranean, and Indian cuisines. No shared rooms for accommodations would be ideal. We do not have preferences for transportation.\\
Steps:\\
\# Destination cities \# \\
\# Execute CitySearch to identify all potential destination cities in Colorado State for the dates ['2022-03-11', '2022-03-12', '2022-03-13', '2022-03-14', '2022-03-15', '2022-03-16', '2022-03-17'] from the starting point 'Indianapolis', and exclude 'Indianapolis' if it appears in the list\\
\# Iterate over cities to select 3 destination cities\\
\# Initialize Z3 solver s\\
\# Define 'city' variable to represent indexes of 3 destination cities\\
\# Ensure city\_0\_index, city\_1\_index, city\_2\_index are distinct, and assert 3 'city' variables equal to city index\\

\# Departure dates \# \\
\# Define 'departure\_dates' variables for 4 transportations between cities\\
\# Assert that the first transportation occurs on the first day (day 0), the last transportation on the last day (day 6), and the second and third transportations occur in between but not on the same day\\

\# Transportation methods \# \\
\# Define transportation method (flight, self-driving, taxi) variable for 4 transportations between cities\\
\# Assert that only one of flight, self-driving, or taxi is used for 4 transportations between cities, and self-driving is invalid if taxi or flight is used for any transportation\\

\# Flight information \# \\
\# Execute FlightSearch to obtain flight info for Indianapolis as the origin, list of cities, city\_0, city\_1, and city\_2, and dates ['2022-03-11', '2022-03-12', '2022-03-13', '2022-03-14', '2022-03-15', '2022-03-16', '2022-03-17']\\
\# Retrieve specific flight price info with Indianapolis as the origin and final destination, specific city variable, and departure date for 4 transportations\\
\# Define 'flight\_index' variable for 4 transportations\\
\# Assert that 4 'flight\_index' variables are within a valid range if taking a flight, and set flight index to -1 if not taking a flight\\
\# Calculate flight price for 5 people for 4 transportations based on flight index variable\\
\# Retrieve specific flight arrival time info with Indianapolis as the origin and final destination, specific city, and departure date for 4 transportations\\
\# Calculate flight arrival time for 4 transportations based on flight index variable\\

\# Driving information \# \\
\# Execute DistanceSearch to obtain driving info for Indianapolis as the origin and city\_0, city\_1, and city\_2\\
\# Retrieve specific driving distance info with Indianapolis as the origin and final destination, specific city, and departure date for 4 transportations\\
\# Assert that driving info is not empty if driving\\
\# Calculate self-driving and taxi price for 5 people and 4 transportations based on driving distance\\
\# Retrieve driving arrival time with Indianapolis as the origin and final destination, specific city, and departure date for 4 transportations\\

\# Restaurant information \# \\
\# Obtain arrivals and city list for each day based on 4 transportations, 7 total travel days, and departure dates variables\\
\# Execute RestaurantSearch to obtain restaurant price info and cuisine info for city\_0, city\_1, and city\_2\\
\# Define 'restaurant\_in\_which\_city' variables for 21 (3 meals per day, 7 days) meals\\
\# For each 'restaurant\_in\_which\_city' variable, assert it to be either the current city or the next city based on transportation arrivals time\\
\# Define 'restaurant\_index' variables for 21 (3 meals per day, 7 days) meals\\
\# For each 'restaurant\_index', retrieve specific price info based on 'restaurant\_in\_which\_city' variable, assert indexes are within a valid range, assert restaurants in the same city are not repeated, and calculate restaurant price for 5 people\\
\# Define 'cuisine\_type' variables for each cuisine type ['Mexican', 'Italian', 'Mediterranean', 'Indian'] required\\
\# For each cuisine type, iterate through all restaurants to check if it is satisfied\\

\# Attraction information \# \\
\# Execute AttractionSearch to obtain attraction info for city\_0, city\_1, and city\_2\\
\# Define 'attraction\_in\_which\_city' variables for 7 (1 per day) attractions\\
\# For each 'attraction\_in\_which\_city' variable, assert it to be either the current city or the next city based on transportation arrivals time\\
\# Define 'attraction\_index' variables for 7 (1 per day) attractions\\
\# For each 'attraction\_index', retrieve specific length info based on attraction in which city variable, assert indexes are within a valid range, and attractions in the same city are not repeated\\

\# Accommodation information \# \\
\# Execute AccommodationSearch to obtain accommodation info and accommodation constraints for city\_0, city\_1, and city\_2\\
\# Define 'accommodation\_index' variables for 3 (1 per city) accommodations\\
\# For each 'accommodation\_index', retrieve specific price info based on accommodation in which city variable, assert 'accommodation\_index' variable is within a valid range, calculate the number of rooms needed for 5 people and accommodation price\\
\# For each city, obtain accommodation minimum night info and assert it to be less than the days stayed in this city\\
\# For each 'accommodation\_index', retrieve specific room type and house rules info, assert 'Shared room' does not exist for all accommodations, assert 'No pets' does not exist for all accommodations, assert 'No children under 10' does not exist for all accommodations\\

\# Budget \# \\
\# Set budget limit variable to be 15100\\
\# Add 4 transportation prices to spent, according to whether the transportation method is flight, self-driving, or taxi\\
\# Add restaurant price to spent\\
\# Add accommodation price to spent\\
\# Assert current spent is within budget\\

-----EXAMPLE 3-----\\
Natural Language query:\\
Please create a 3-day travel itinerary for 2 people beginning in Fort Lauderdale and ending in Milwaukee from the 8th to the 10th of March, 2022. Our travel budget is set at \$1,100. We'd love to experience both American and Chinese cuisines during our journey. We'd love to live in private rooms but we don't want to take flights. \\
Steps:\\
\# Destination cities \# \\
\# Set cities to be a list that includes Milwaukee only\\
\# Iterate over cities for 1 destination city\\
\# Initialize Z3 solver s\\
\# Define 'city' variable to represent indexes of 1 destination city\\
\# Assert 'city' variable equals city index\\

\# Departure dates \# \\
\# Define 'departure\_dates' variables for 2 transportations between cities\\
\# Assert that the first transportation occurs on the first day (day 0), and the last transportation on the last day (day 2)\\

\# Transportation methods \# \\
\# Define transportation method (flight, self-driving, taxi) variable for 2 transportations between cities\\
\# Assert that only one of flight, self-driving, or taxi is used for 2 transportations between cities, and self-driving is invalid if taxi or flight is used for any transportation\\
\# Assert that all 2 transportations between cities are not flights\\

\# Flight information \# \\
\# Execute FlightSearch to obtain flight info for Fort Lauderdale as the origin, list of cities, city\_0, and dates ["2022-03-08", "2022-03-09", "2022-03-10"]\\
\# Retrieve specific flight price info with Fort Lauderdale as the origin and final destination, specific city, and departure date for 2 transportations\\
\# Define 'flight\_index' variable for 2 transportations\\
\# Assert that 2 'flight\_index' variables are within a valid range if taking a flight, and set flight index to -1 if not taking a flight\\
\# Calculate flight price for 2 people for 2 transportations based on flight index variable\\
\# Retrieve specific flight arrival time info with Fort Lauderdale as the origin and final destination, specific city, and departure date for 2 transportations\\
\# Calculate flight arrival time for 2 transportations based on flight index variable\\

\# Driving information \# \\
\# Execute DistanceSearch to obtain driving info for Fort Lauderdale as the origin and city\_0\\
\# Retrieve specific driving distance info with Fort Lauderdale as the origin and final destination, specific city, and departure date for 2 transportations\\
\# Assert that driving info is not empty if driving\\
\# Calculate self-driving and taxi price for 2 people and 2 transportations based on driving distance\\
\# Retrieve driving arrival time with Fort Lauderdale as the origin and final destination, specific city, and departure date for 2 transportations\\

\# Restaurant information \# \\
\# Obtain arrivals and city list for each day based on 2 transportations, 3 total travel days, and departure dates variables\\
\# Execute RestaurantSearch to obtain restaurant price info and cuisine info for city\_0\\
\# Define 'restaurant\_in\_which\_city' variables for 9 (3 meals per day, 3 days) meals\\
\# For each 'restaurant\_in\_which\_city' variable, assert it to be either the current city or the next city based on transportation arrivals time\\
\# Define 'restaurant\_index' variables for 9 (3 meals per day, 3 days) meals\\
\# For each 'restaurant\_index', retrieve specific price info based on 'restaurant\_in\_which\_city' variable, assert indexes are within a valid range, assert restaurants in the same city are not repeated, and calculate restaurant price for 2 people\\
\# Define 'cuisine\_type' variables for each cuisine type ["American", "Chinese"] required\\
\# For each cuisine type, iterate through all restaurants to check if it is satisfied\\

\# Attraction information \# \\
\# Execute AttractionSearch to obtain attraction info for city\_0\\
\# Define 'attraction\_in\_which\_city' variables for 3 (1 per day) attractions\\
\# For each 'attraction\_in\_which\_city' variable, assert it to be either the current city or the next city based on transportation arrivals time\\
\# Define 'attraction\_index' variables for 3 (1 per day) attractions\\
\# For each 'attraction\_index', retrieve specific length info based on attraction in which city variable, assert indexes are within a valid range, and attractions in the same city are not repeated\\

\# Accommodation information \# \\
\# Execute AccommodationSearch to obtain accommodation info and accommodation constraints for city\_0\\
\# Define 'accommodation\_index' variables for 1 (1 per city) accommodation\\
\# For each 'accommodation\_index', retrieve specific price info based on accommodation in which city variable, assert 'accommodation\_index' variable is within a valid range, calculate the number of rooms needed for 2 people and accommodation price\\
\# For each city, obtain accommodation minimum night info and assert it to be less than the days stayed in this city\\
\# For each 'accommodation\_index', retrieve specific room type info, assert 'Private room' exists for all accommodations\\

\# Budget \# \\
\# Set budget limit variable to be 1100\\
\# Add 2 transportation prices to spent, according to whether the transportation method is flight, self-driving, or taxi\\
\# Add restaurant price to spent\\
\# Add accommodation price to spent\\
\# Assert current spent is within budget\\

-----EXAMPLES END-----\\

Using the examples provided, outline the steps for the following Natural Language query. Adhere to the original step structures. Begin each line containing steps with '\#'. 

Natural Language query:
\end{boxC}

Here's an example paraphrased prompt for Step-Code (Destination City constraint type):
\begin{minted}[fontsize=\scriptsize,breaklines]{python}
# Python script to evaluate the satisfiability of constraints related to destination cities in a travel plan problem.

# Execute CitySearch to retrieve all potential destination cities in Minnesota for the dates ["2022-03-16", "2022-03-17", "2022-03-18"], starting from 'Atlanta'. Exclude 'Atlanta' from the list if present.
cities = CitySearch.run('Minnesota', 'Atlanta', ["2022-03-16", "2022-03-17", "2022-03-18"])
if 'Atlanta' in cities:
    cities.remove('Atlanta')
# Assign cities to a list containing only 'Milwaukee'
cities = ['Milwaukee']
# Iterate over cities to select 2 destination cities
for city_0_index, city_0 in enumerate(cities):
    for city_1_index, city_1 in enumerate(cities):
        # Create a Z3 solver instance named s
        s = Optimize()
        # Define 'city' variable as indices of 2 destination cities
        variables['city'] = [Int('city_' + str(i)) for i in range(2)]
        # If city_0_index and city_1_index differ, assert that the 2 'city' variables match the city indices
        if city_0_index != city_1_index:
            s.assert_and_track(variables['city'][0] == city_0_index, 'visit city in cities list')
            s.assert_and_track(variables['city'][1] == city_1_index, 'visit city in cities list')
# Iterate over cities to select 3 destination cities
for city_0_index, city_0 in enumerate(cities):
    for city_1_index, city_1 in enumerate(cities):
        for city_2_index, city_2 in enumerate(cities):
            # Create a Z3 solver instance named s
            s = Optimize()
            # Define 'city' variable as indices of 3 destination cities
            variables['city'] = [Int('city_' + str(i)) for i in range(3)]
            # If city_0_index, city_1_index, and city_2_index are distinct, assert that the 3 'city' variables match the city indices
            if city_0_index != city_1_index and city_1_index != city_2_index and city_0_index != city_2_index:
                s.assert_and_track(variables['city'][0] == city_0_index, 'visit city in cities list')
                s.assert_and_track(variables['city'][1] == city_1_index, 'visit city in cities list')
                s.assert_and_track(variables['city'][2] == city_2_index, 'visit city in cities list')
# Iterate over cities to select 1 destination city
for city_0_index, city_0 in enumerate(cities):
    # Create a Z3 solver instance named s
    s = Optimize()
    # Define 'city' variable as the index of 1 destination city
    variables['city'] = [Int('city_' + str(i)) for i in range(1)]
    # Assert that the 'city' variable matches the city index
    s.assert_and_track(variables['city'][0] == city_0_index, 'visit city in cities list')

# From the examples provided, lines beginning with '#' are instructions, and the lines following them until the next '#' are the corresponding code.
# Adhere to the code structure and variable names used in the examples.
# For the instruction below, write the corresponding code and respond with the code only. Begin with ########## Destination cities response########## and conclude with ########## Destination cities response ends##########.
\end{minted}
\clearpage
\section{Example input query and output step for generalization evaluation}
\label{sec:gen-ex}
We include the full example of Query-Step generation for unseen constraint types here:\\ 
\begin{boxC}
\textbf{Input JSON query}
\{\\
    "org": "Hong Kong", \\
    "dest": ["New York City", "Bangkok"], \\
    "days": 5, \\
    "visiting\_city\_number": 2, \\
    "date": ["2023-12-22", "2023-12-23", "2023-12-24", "2023-12-25", "2023-12-26"], \\
    "people\_number": 3, \\
    "local\_constraint": \{\\
        "flight rule": "non-stop", \\
        "airlines": ["United", "Emirates"], \\
        "attraction\_category": ["Garden", "Historical Landmarks"],\\
        "transportation": "flight"\\
    \}, \\
    "budget": 5000\\
\}\\
\textbf{Corresponding output step}
\# Destination cities \# \\
\# Set cities to be a list includes 'New York City' and 'Bangkok'\\
\# Loop through cities for 2 destination cities\\
\# Initialize Z3 solver s\\
\# Set 'city' variable to be indexes of 2 destination cities\\
\# If city\_0\_index and city\_1\_index are not same, assert 2 'city' variables equal to city index\\
\\
\# Departure dates \# \\
\# Set 'departure\_dates' variables for 3 transportations between cities\\
\# Assert first transportation happens at first day (day 0), last transportation happens at last day (day 4), and second transportation could happen at any day in between\\

\# Transportation methods \# \\
\# Set transportation method (flight) variable for 3 transportations between cities\\
\# Assert all 3 transportations between cities are flight\\

\# Flight information \# \\
\# Run FlightSearch to get flight info for Hong Kong as origin, city\_0 and city\_1, and dates\\
\# Get specific flight price info with Hong Kong as origin and final destination, specific city variable, and departure date for 3 transportations\\
\# Set 'flight\_index' variable for 3 transportations\\
\# Assert 3 'flight\_index' variables are within valid range if taking flight, assert flight index to be -1 if not taking flight\\
\# Calculate flight price for 3 people for 3 transportations based on flight index variable\\
\# Get specific flight arrival time info with Hong Kong as origin and final destination, specific city, and departure date for 3 transportations\\
\# Calculate flight arrival time for 3 transportations based on flight index variable\\
\# For each 'flight\_index', get specific flight rule and airlines info, assert 'non-stop' exist for all flights, assert 'United' or 'Emirates' exist for all flights\\
\\
\# Attraction information \# \\
\# Get arrivals and city list for each day based on 3 transportations, 5 total travel day, and departure dates variables\\
\# Run AttractionSearch to get attraction info for city\_0 and city\_1\\
\# Set 'attraction\_in\_which\_city' variables for 5 (1 per day) attractions\\
\# For each 'attraction\_in\_which\_city' variable, assert it to be either current city or next city based on transportation arrivals time\\
\# Set 'attraction\_index' variables for 5 (1 per day) attractions\\
\# For each 'attraction\_index', get specific length info based on attraction in which city variable, assert index are within valid range, and attrations in same city are not repeated\\
\# Set 'attraction\_category' variables for each category required\\
\# For each category, iterate through all attraction to check if it is satisfied\\
\\
\# Restaurant information \# \\
\# Set restaurant price to be 30 per person per meal, calculate total restaurant price for 3 people for 5 days\\
\\
\# Accommodation information \# \\
\# Set accommodation price to be 100 per person per night, calculate total accommodation price for 3 people for 4 nights\\
\\
\# Budget \# \\
\# Set budget limit variable to be 5000\\
\# Add 3 transportation price to spent, according to whether transportation method is flight\\
\# Add restaurant price to spent\\
\# Add accommodation price to spent\\
\# Assert current spent is within budget\\
\end{boxC}

\end{document}